\documentclass[preprint,12pt]{elsarticle}

\usepackage{amssymb}

\usepackage{amsmath}
\usepackage{xcolor}
\usepackage{hyperref}
\usepackage{gensymb}
\usepackage{pifont}
\newcommand{\cmark}{\ding{51}}
\newcommand{\xmark}{\ding{55}}
\usepackage{wrapfig}

\journal{}

\begin{document}

\begin{frontmatter}



\title{SegForestNet: Spatial-Partitioning-Based Aerial Image Segmentation}


\author[label1]{Daniel Gritzner\corref{cor1}}
\ead{gritzner@tnt.uni-hannover.de}
\cortext[cor1]{corresponding author}

\author[label1]{Jörn Ostermann}
\ead{ostermann@tnt.uni-hannover.de}

\affiliation[label1]{organization={Leibniz Universität Hannover, Fakultät für Elektrotechnik und Informatik, Institut für Informationsverarbeitung},
            addressline={Appelstr. 9A}, 
            city={Hannover},
            postcode={30161}, 
            state={Lower Saxony},
            country={Germany}}

\begin{abstract}
Aerial image segmentation is the basis for applications such as automatically creating maps or tracking deforestation. In true orthophotos, which are often used in these applications, many objects and regions can be approximated well by polygons. However, this fact is rarely exploited by state-of-the-art semantic segmentation models. Instead, most models allow unnecessary degrees of freedom in their predictions by allowing arbitrary region shapes. We therefore present a refinement of our deep learning model which predicts binary space partitioning trees, an efficient polygon representation. The refinements include a new feature decoder architecture and a new differentiable BSP tree renderer which both avoid vanishing gradients. Additionally, we designed a novel loss function specifically designed to improve the spatial partitioning defined by the predicted trees. Furthermore, our expanded model can predict multiple trees at once and thus can predict class-specific segmentations. As an additional contribution, we investigate the impact of a non-optimal training process in comparison to an optimized training process. While model architectures optimized for aerial images, such as PFNet or our own model, show an advantage under non-optimal conditions, this advantage disappears under optimal training conditions. Despite this observation, our model still makes better predictions for small rectangular objects, e.g., cars.
\end{abstract}




\end{frontmatter}


\section{Introduction}
\label{sec:introduction}

\newcommand{\todo}[1]{\textcolor{red}{\textbf{TODO:}} {#1}}

Computer vision techniques, such as object detection and semantic segmentation, are used for many aerial image analysis applications. As an example, traffic can be analyzed by using object detection to find vehicles in aerial images. Semantic segmentation is the basis for creating and updating maps \cite{girard2018aligning} and can be used for tracking city growth \cite{gomez2020spatiotemporal} or tracking deforestation \cite{lee2020classification}.

However, any error in the prediction of a computer vision model, i.e., object detection or semantic segmentation, causes errors in applications based on said prediction. Precise predictions are required, which can be achieved by exploiting domain knowledge. There are several models which attempt to exploit this knowledge in the domain of aerial image segmentation, e.g., RA-FCN by Mou et al. \cite{mou2019relation}, PFNet by Li et al. \cite{li2021pointflow} and our own BSPSegNet \cite{gritzner2021semantic}. RA-FCN explicitly models long-range spatial and channel relations, PFNet performs point-wise affinity propagation to mitigate class imbalances and BSPSegNet predicts a limited number of polygons in order to approximate segments without unnecessary degrees of freedom, especially in true orthophotos.

\begin{figure}[t]
	\begin{center}
		\includegraphics[width=\columnwidth]{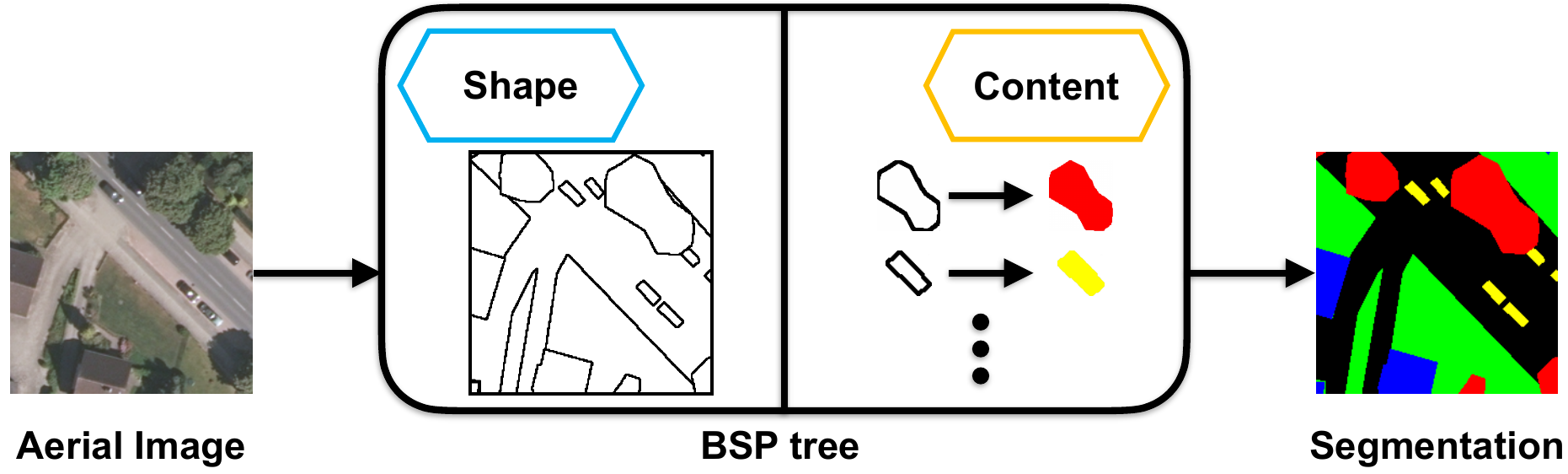}
	\end{center}
	\caption{Overview of our approach: our model predicts binary space partitioning (BSP) trees from aerial images. The inner nodes of such a tree define the shape of regions, while the leaf nodes define the content of each region. The leaf nodes effectively map shapes to classes. BSP trees can be rendered in a differentiable way into a full segmentation thus enabling end-to-end model training. Our proposed refinements improve gradient computations in the BSP renderer and parts of the model. Additionally, a novel loss function improves the predicted inner node parameters, i.e., the predicted shapes. Furthermore, we extended the approach to predict multiple trees at the same time in order to enable class-specific shape predictions.}
	\label{fig:teaser}
\end{figure}

We created BSPSegNet, a model for aerial image segmentation which predicts polygons as shown in Fig. \ref{fig:teaser}.  The model partitions the input image into $8\times 8$ blocks. For each block, a binary space partitioning (BSP) tree \cite{fuchs1980visible} with depth 2 is predicted, i.e., every block is partitioned into up to four segments with a separate class prediction for each segment. The parameters of the inner nodes of the BSP tree encode parameters of a line partitioning a region into two subregions, while the parameters of the leaf nodes are class logits. The lines encoded by the inner nodes together with the block boundaries form polygons. As we demonstrated in our earlier paper, the ground truth of common aerial image datasets can be encoded extremely accurately this way ($99\%$ accuracy and $99\%$ mean Intersection-over-Union (mIoU)), despite arbitrary region shapes no longer being feasible predictions. However, arbitrarily shaped regions, e.g., circle-shaped cards, would not be meaningful anyway. Therefore, limiting the prediction budget by limiting the number of predicted polygons serves as a helpful inductive bias for the model.

While BSPSegNet already performed well, we present several improvements to its performance in this article. While investigating ways to improve BSPSegNet, we also improved our model training process which of course also improved the performance of other models in our comparison as well. This lead to an interesting observation: a well-optimized training process is more important for aerial image segmentation than exploiting domain knowledge despite claims made by authors of aerial image optimized model architectures. With the proper training process, many state-of-the-art models perform equally well, including models not specifically optimized for aerial images. Therefore, our contributions in this article are two-fold: we compare several state-of-the-art models using two different training processes and we introduce SegForestNet, an improved BSPSegNet. Our contributions are:
\begin{enumerate}
    \item We replace BSPSegNet's two feature decoders with an architecture using residual connections and modify the model's differentiable BSP tree renderer to avoid using sigmoid activations. BSPSegNet tries to push the output of these activations close to $0$ or $1$. However, in these value regions the derivative of the sigmoid function is almost $0$. Vanishing gradients are avoided due to these modifications.
    \item We introduce a novel loss function punishing BSP trees which create partitions containing multiple different classes (according to the ground truth). This loss specifically affects only the inner nodes of the BSP trees, instead of all the nodes as categorical cross entropy does, thus improving the shapes of the predicted partitions.
    \item We expand BSPSegNet's prediction s.t. it is able to predict multiple BSP trees for each $8\times 8$ block, thus enabling class-specific segmentations for more precise shape predictions.
    \item We compare two different model training processes which use different hyperparameter settings, e.g., altering how augmentation is applied during training.
\end{enumerate}
With these modifications SegForestNet can be trained end-to-end without the two phase training previously recommended in \cite{gritzner2021semantic}. In the first phase, an autoencoder phase, a mapping of the ground truth to BSP trees was learned. In the second phase, this new ground truth representation was used to learn the actual semantic segmentation without using the differentiable BSP renderer and thus avoiding its effects on gradients. Our modifications allow the faster and more efficient training of SegForestNet without the autoencoder phase.

In addition to the above contributions, we also experimented with different signed distance functions (SDFs) and tree structures. SDFs are used by the inner nodes to partition a segment into subsegments. By trying different SDFs, we were able to generalize the shape of the partitioning boundary from lines to other shapes such as circles. Also, our approach generalizes to different tree structures such as quadtrees \cite{finkel1974quad} and k-d trees \cite{bentley1975multidimensional}. However, we found that neither of these two changes provide any benefit for the semantic segmentation of aerial images. Since we believe that these ideas may be useful in other domains with more organic shapes, e.g., the segmentation of images captured by cameras of self-driving cars, we show how to generalize SegForestNet to different SDFs and tree structures in \ref{app:generalization}.

In this article's next section we will discuss related work, followed by a description of SegForestNet and how we trained the models for our comparison. We then evaluate our contributions by comparing SegForestNet on several aerial image datasets to multiple state-of-the-art semantic segmentation models using two different training process configurations. We finally conclude with a summary.

Our implementation can be found on GitHub\footnote{\url{https://github.com/gritzner/SegForestNet}}.

\section{Related Work}
Modern semantic segmentation models rely on deep learning and an encoder-decoder structure \cite{papandreou1502weakly,yu2015multi,jegou2017one,chen2017rethinking,wu2019fastfcn,PesYan2022}. An encoder is used to map an input image to a feature map with a reduced spatial resolution but a much higher number of channels/features. While a large number of features is computed, the spatial information is bottlenecked. This implies that downsampling, e.g., through max-pooling or strided convolutions, is performed. Using such a feature map, a decoder predicts the desired semantic segmentation. To get a segmentation of the same resolution as the input image, the decoder uses bilinear sampling and/or transpose convolutions (also called deconvolutions) in order to upsample the spatial resolution of the feature map back to the initial input resolution.

U-Net \cite{ronneberger2015u} is a popular model following this paradigm. Though designed for medical image analysis, it is also used for other types of data, including even 3D data \cite{cciccek20163d}. U-Net's encoder and decoder are almost symmetrical and skip connections at each spatial resolution are used to allow predictions with high frequency details. Other models, such as Fully Convolutional Networks for Semantic Segmentation (FCN) \cite{long2015fully}, have an asymmetrical architecture with a significantly more sophisticated encoder than decoder.
They rely on existing classification models, such as VGG \cite{simonyan2014very}, ResNet \cite{he2016deep}, MobileNetv2 \cite{sandler2018mobilenetv2}, Xception 
\cite{chollet2017xception}, or ConvNeXt \cite{liu2022convnet} as encoder (also called backbone or feature extractors in these cases) by removing their final classification layers.
Instead, semantic segmentation models append a novel decoder to these encoders. The complexity of these decoders varies. While FCN uses a rather simple decoder, DeepLab v3+ \cite{chen2018encoder} uses a more sophisticated decoder and also adds atrous spatial pyramid pooling \cite{chen2017deeplab} to the encoder. By doing so, DeepLab v3+ is able to account for context information at different scales. Other modifications to the basic encoder-decoder idea are adding parallel branches to process specific features such as edges/boundaries \cite{takikawa2019gated}, processing videos instead of images \cite{zhu2019improving}, explicitly modelling long-range spatial and channel relations \cite{mou2019relation}, separating the image into fore- and background \cite{zheng2020foreground}, or point-wise affinity propagation to handle imbalanced class distributions with an unusually high number of background pixels \cite{li2021pointflow}. However, not all modern segmentation models use complex decoders. SegNeXt \cite{guo2022segnext} uses a simple decoder similar to FCN's decoder but adds a low rank embedding just prior to the final classification layer. Vision transformer-based models \cite{thisanke2023semantic}, inspired by the transformers used for large language models \cite{vaswani2017attention}, have recently become popular  even though other recent publications using convolutional models \cite{liu2022convnet,guo2022segnext} demonstrated that transformer-based models do not inherently offer better performance for vision tasks such as classification or semantic segmentation.

More recent research also works on more complex problems, e.g., instance segmentation  \cite{he2017mask} and panoptic segmentation \cite{kirillov2019panoptic,xiong2019upsnet,yang2019deeperlab,yang2020sognet}. In instance segmentation pixel masks of detected objects, such as cars or persons, are predicted. Panoptic segmentation is a combination of semantic segmentation and instance segmentation: a class is predicted for each pixel in the input image while for all pixels of countable objects an additional instance ID is predicted. Models dealing with these more complex tasks still usually require a semantic segmentation as a component to compute the final prediction. AdaptIS \cite{sofiiuk2019adaptis} is such a model. When performing instance segmentation, it uses U-Net as a backbone while using ResNet and DeepLab v3+ when performing panoptic segmentation.

Our previous model, BSPSegNet \cite{gritzner2021semantic}, predicts BSP trees to perform a semantic segmentation. These trees are usually used in computer graphics \cite{sanglard2019game} rather than computer vision. Each inner node of a BSP tree partitions a given region into two segments while the leaf nodes define each segment's content, i.e., its class in the case of a semantic segmentation. Since BSP trees are resolution-indepedent, BSPSegNet's decoder does not have to perform any kind of upsampling. Furthermore, BSPSegNet is the only segmentation model which inherently semantically separates features into shape and content features. Similar works using computer graphics techniques for computer vision are
PointRend \cite{kirillov2020pointrend} and 
models reconstructing 3D meshes 
\cite{tatarchenko2017octree,wang2018pixel2mesh,mescheder2019occupancy,gkioxari2019mesh}. PointRend iteratively refines a coarse instance segmentation into a finer representation. BSPnet \cite{chen2020bsp} also uses BSP, but instead of a segmentation, it tries to reconstruct polygonal 3D models. Additionally, rather than predicting a BSP tree hierarchy, it instead predicts a set of BSP planes.

As mentioned before, BSP trees are a data structure more common in computer graphics than in computer vision. Other commonly used data structures in computer graphics and other applications, e.g., data compression, are $k$-d trees \cite{bentley1975multidimensional}, quadtrees \cite{finkel1974quad}, and bounding volume hierarchies \cite{ericson2004real}. They differ from BSP trees in that they perform a partitioning using an axis-aligned structure which makes slopes expensive to encode compared to BSP trees.

\section{Semantic Segmentation through Spatial Partitioning}
\label{sec:method}

We will first briefly describe BSPSegNet's architecture in this section in order to be able to precisely discuss our contributions. The third subsection presents the first contribution mentioned in the introduction, which reduces the risk of vanishing gradients, while the fourth subsection presents our second contribution, the novel loss function. The fifth subsection then discusses the changes necessary to enable to the prediction of multiple partitioning trees simultaneously, our third contribution. In the final subsection we give an overview of the model training processes we used in our evaluation. The actual hyperparameter settings for both processes used for our fourth contribution can be found in subsection \ref{sec:trainingprocessvariants}.

\subsection{BSPSegNet}
\begin{figure}
	\begin{center}
		\includegraphics[width=\columnwidth]{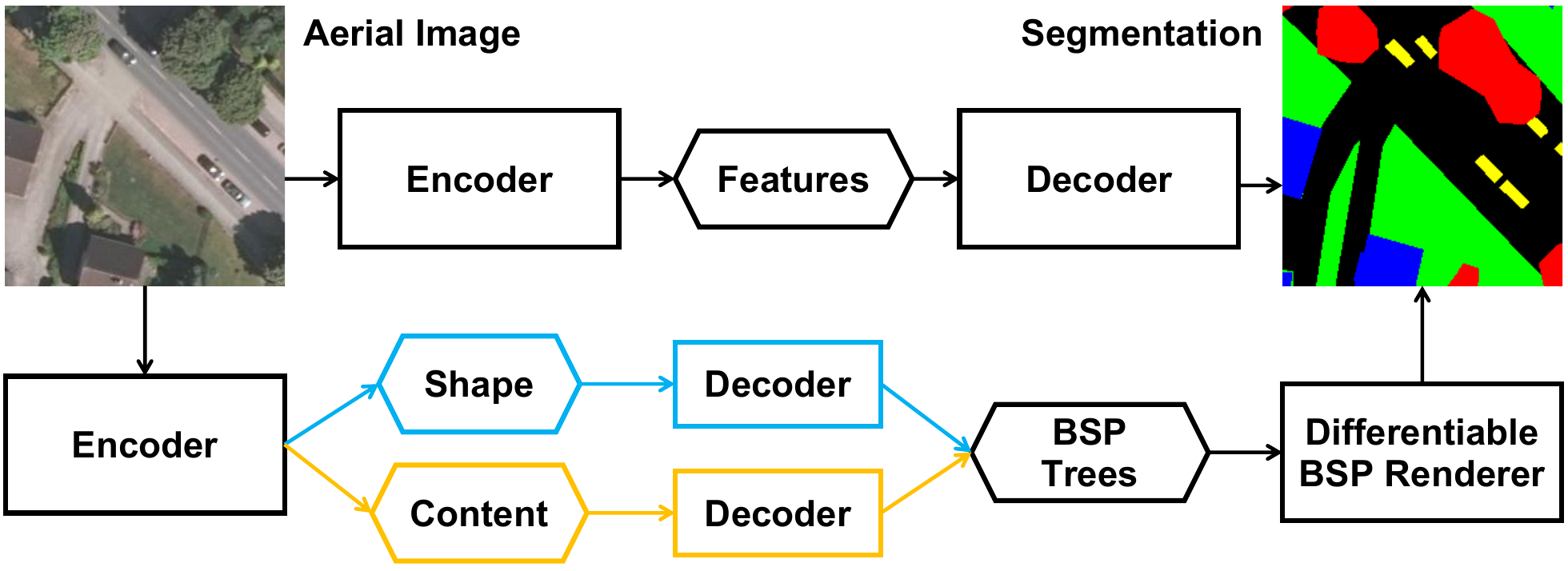}
	\end{center}
	\caption{A comparison of state-of-the-art models (top) and BSPSegNet/SegForestNet (bottom; ours). All models use an encoder-decoder architecture, however, our models semantically splits the feature map (hexagon) into shape and content features. These are decoded into the inner nodes and leaf nodes of BSP trees respectively (see Fig. \ref{fig:bsptree}). Our models also needs a differentiable BSP renderer to enable end-to-end training. The renderer is a fixed function without learnable parameters.}
	\label{fig:model}
\end{figure}

Fig. \ref{fig:model} (top) shows a common encoder-decoder architecture used by state-of-the-art models. An encoder maps an input image to a feature map with a bottlenecked spatial resolution. This feature map is then decoded into a semantic segmentation at the same spatial resolution as the input image. Skip connections at different spatial resolutions help to maintain high frequency details.

\begin{figure}
	\begin{center}
		\includegraphics[width=\columnwidth]{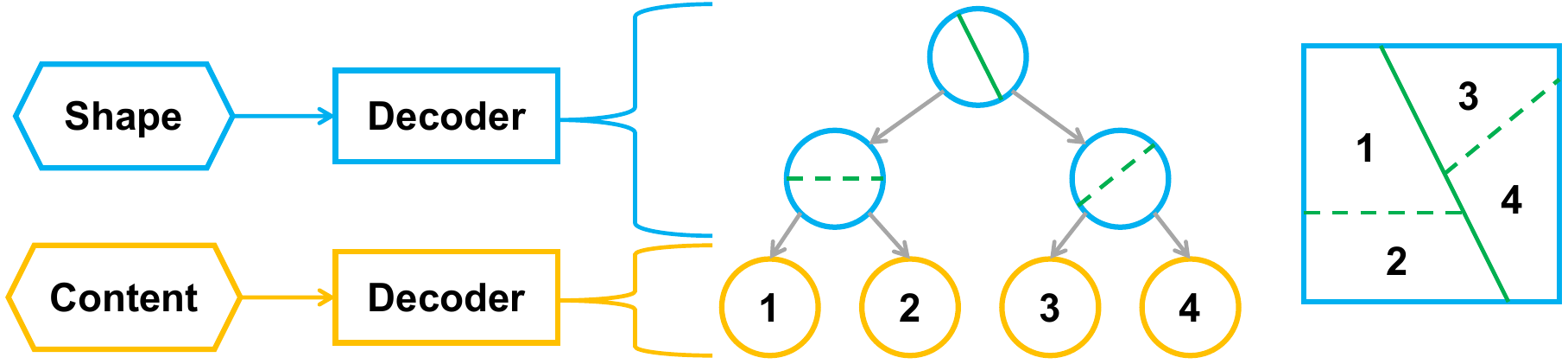}
	\end{center}
	\caption{A partitioning of a square region (right) defined by a BSP tree (center). Shape features are decoded 
	into the parameters of the inner nodes (blue), which define lines (green) creating the partitioning. Content features are decoded into the parameters of the leaf nodes (orange), which are the class logits predicted for each partition.}
	\label{fig:bsptree}
\end{figure}

The bottom of Fig. \ref{fig:model} shows BSPSegNet in comparison. As other state-of-the-art models, it uses an encoder-decoder architecture. However, it uses two separate decoders. The feature map computed by the encoder is split along the feature dimension into two separate maps: one for shape features (blue) and one for content features (orange). Each feature map is processed by a different decoder. While the decoders share the same architecture, each has its own unique weights. The shape features are decoded into the inner nodes of the BSP trees, which represent the shape of the regions in the final segmentation, as shown in Fig. \ref{fig:bsptree}. The content features are decoded into the leaf nodes of the BSP trees, which represent each region's class logits.

Like other semantic segmenation models, BSPSegNet uses the feature extraction part of a classification model like MobileNetv2 or Xception as an encoder. By removing pooling operations or reducing the stride of convolution operations, the encoder is modified such that the input image is downsampled by a factor of $8$ along each spatial dimension. To keep in line with classification model design, the earlier downsampling operations are kept while the later ones are removed. This enables more efficient inference and training, as the spatial resolution is reduced early. Furthermore, the number of features in  the final feature map computed by the encoder is reduced such that there is a bottleneck.
The shape feature map has fewer features than the number of parameters required for the inner nodes, while the content feature map has fewer features than the parameters required for the leaf nodes.

BSPSegNet subdivides the input image into $8\times 8$ blocks (the same as the downsampling factor in the encoder) and predicts a separate BSP tree for each block, i.e., the spatial resolution of the feature map is such that there is exactly one spatial unit per block. The two decoders keep the spatial resolution and only modify the number of features. This is done by concatenating three blocks, each consisting of a $1\times 1$ convolution outputting $256$ features, batch normalization and a LeakyReLU activation. These blocks are followed by one final $1\times 1$ convolution to predict the final BSP tree parameters.

\subsection{BSPSegNet's Differentiable BSP Renderer}
The differentiable BSP renderer is based on the idea of a region map $\textbf{R}$. For each pixel $\textbf{p} = (x,y)$, the entry $\textbf{R}_{iyx}$ is the probability that $\textbf{p}$ is part of the $i$-th region (leaf node). This implies $\sum_{i=1}^k 
\textbf{R}_{iyx} = 1$ for all pixels $\textbf{p}$ with $k$ equal to the number of leaf nodes.

Initially, BSPSegNet starts with all $\textbf{R}_{iyx}$ set to $1$ and then updates the region map for every inner node. Each inner node consists of three parameters, two parameters for the normal vector $\textbf{n}$ and one parameter $d$. These parameters together define a line in the two-dimensional image space. BSPSegNet does neither normalize $\textbf{n}$, nor does it enforce $|\textbf{n}| = 1$ through any kind of loss. Rather, it lets the network learn to predict an appropriately scaled $d$. For every inner node and pixel $\textbf{p}$, the signed distance function
$ f(\textbf{p}) = \textbf{n}\cdot\textbf{p} - d $
is computed. The equations
\begin{equation}
    g(\textbf{p}) = \sigma(\lambda_1 \cdot f(\textbf{p}))
    \label{eq:sigmoid}
\end{equation}
\begin{equation}
    \textbf{R}_{iyx} := \textbf{R}_{iyx} \cdot \lambda_2 \cdot g(\textbf{p})
    \label{eq:rleft}
\end{equation}
\begin{equation}
    \textbf{R}_{iyx} := \textbf{R}_{iyx} \cdot \lambda_2 \cdot (1 - g(\textbf{p}))
    \label{eq:rright}
\end{equation}
with the sigmoid function $\sigma$ are then used to update the region map $\textbf{R}$. For each inner node, equation \ref{eq:rleft} is used for all leaf nodes $i$ reachable from the left child node while equation \ref{eq:rright} is used for all $i$ reachable by the right child node.

Depending on where the pixel $\textbf{p}$ lies, the function $g$ either approaches $0$ or $1$. It approaches $1$ if $\textbf{p}$ is reachable via the left child node and it approaches $0$ if $\textbf{p}$ is reachable via the right child node. Therefore, $\textbf{R}_{iyx}$ is multiplied with a value close to $0$ for those regions $i$ which $\textbf{p}$ does not belong to. Thus, eventually, all $\textbf{R}_{iyx}$ for $i$ to which $\textbf{p}$ does not belong will be close to $0$. The one remaining entry $\textbf{R}_{iyx}$ will be close to $\lambda_2^D$, where $D$ is the depth of the BSP tree.

After iterating over all inner nodes, a Softmax operation across the region dimension $i$ is applied. This ensures that each $\textbf{R}_{iyx}$ represents the aforementioned probability. The hyperparameter $\lambda_2$ controls how close the respective final entries of $\textbf{R}$ will be to $0$ or $1$, while $\lambda_1$ controls how large $|f(\textbf{p})|$ must be in order for $\textbf{p}$ to be assigned to one specific child node ($g(\textbf{p})$ close to $0$ or $1$) instead of the model expressing uncertainty about which child region $\textbf{p}$ belongs to ($g(\textbf{p}) \approx 0.5$).

The final output $h$ of the BSP renderer is
\begin{equation}
    h(\textbf{p}) = \sum_{i=1}^k \textbf{R}_{iyx} \cdot \textbf{v}_i
    \label{eq:finalprediction}
\end{equation}
with the vectors $\textbf{v}_i$ of class logits, one vector for each leaf node. The function $h$ can be computed in parallel for all pixels $\textbf{p}$. BSP trees are resolution-independent, therefore the final output resolution is determined by how densely $h$ is sampled, i.e., for how many pixels $\textbf{p}$ the function is computed. This is also the reason why BSPSegNet, in contrast to other segmentation models, does not have to perform any kind of upsampling in its decoders. By sampling $h$ at the same resolution as the input image, a pixelwise semantic segmentation can be computed like for any other state-of-the-art model.

\subsection{Refined Gradients}

As mentioned in the introduction, a two-phase training process is recommended for the original BSPSegNet. In the second phase, the differentiable BSP renderer is skipped altogether. We hypothesize that insufficient gradients are responsible: by pushing the computation of $g$ (Eq. \ref{eq:sigmoid}) into value ranges where its derivative is almost $0$ gradients vanish. Additionally, the two feature decoders do not use residual connections, which would help gradients to propagate through the model. Our first contribution aims to improve on these two points in order to avoid the two-phase training process and to enable the use of decoders with more layers which can learn more complex mapping functions from features to BSP tree parameters. The goal is to enable true end-to-end training without the need to learn a mapping to an intermediate representation. This way, training is faster, allowing for more iterations during development, e.g., for hyperparameter optimization, or saving energy and thus reducing costs.

\begin{figure}
	\begin{center}
		\includegraphics[width=\columnwidth]{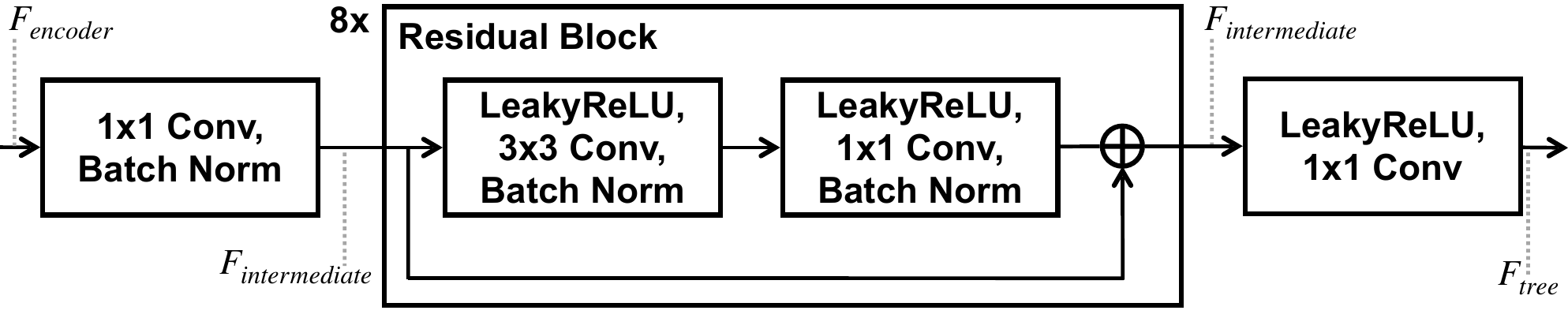}
	\end{center}
	\caption{Our refined decoder architecture using residual blocks. The initial and the last convolution change the number of features $F$. The number of features are set such that there is a bottleneck with regard to the tree parameters, i.e., $F_{encoder} < F_{tree}$.}
	\label{fig:decoder}
\end{figure}

The first part of this contribution is a new feature decoder architecture. Again, we use the same architecture for both decoders, but they still do not share weights. Our new architecture, shown in Fig. \ref{fig:decoder}, starts and ends with a $1\times 1$ convolution, the first one maps the bottlenecked feature dimension coming from the encoder to an intermediate number of features, while the last convolution maps the intermediate features to the final number of tree parameters. In between these two convolutions are several residual blocks. Each block consists of a depthwise $3\times 3$ convolution with zero padding and a $1\times 1$ convolution. The number of features stays constant throughout the residual blocks. Each convolution in the decoders, except for the very last one mapping to the tree parameters, is followed by a batch normalization layer \cite{ioffe2015batch} and a LeakyReLU activation \cite{xu2015empirical}.
This architecture not only uses residual connections for improved gradient propagation, but it also includes spatial context for each block, i.e., for each BSP tree, instead of solely relying on the receptive field of the encoder.

The second part of our first contribution is an updated region map $\textbf{R}$ computation (Eqs. \ref{eq:sigmoid}-\ref{eq:rright}). We start by setting all $\textbf{R}_{iyx}$ to $0$ initially instead of $1$. For each inner node, we use the equations
\begin{equation}
    g(\textbf{p}) = \lambda \cdot f(\textbf{p})
    \label{eq:nosigmoid}
\end{equation}
\begin{equation}
    \textbf{R}_{iyx} := \textbf{R}_{iyx} + \text{ReLU}(g(\textbf{p}))
    \label{eq:rleft2}
\end{equation}
\begin{equation}
    \textbf{R}_{iyx} := \textbf{R}_{iyx} + \text{ReLU}(-g(\textbf{p}))
    \label{eq:rright2}
\end{equation}
for updating $\textbf{R}$ where Eq. \ref{eq:rleft2} replaces Eq. \ref{eq:rleft} and Eq. \ref{eq:rright2} replaces Eq. \ref{eq:rright}. After processing all inner nodes, there still is a Softmax operation across the region dimension $i$. Using our Eqs. \ref{eq:nosigmoid}-\ref{eq:rright2}, we avoid vanishing gradients by replacing the sigmoid function with ReLU. This necessitates moving from multiplications in Eqs. \ref{eq:rleft} and \ref{eq:rright} to additions in Eqs. \ref{eq:rleft2} and \ref{eq:rright2}. As a side effect, the hyperparameters $\lambda_1$ and $\lambda_2$ are merged into just $\lambda$, slightly simplifying the model.

\subsection{Region Map-Specific Loss}

Cross-entropy, the most common loss function used for semantic segmentation, is applied pixelwise and therefore gives the model to be trained a signal for each mispredicted pixel. However, from the point-of-view of BSPSegNet and SegForestNet, there is no distinction between wrong class logits in a leaf node and wrong line parameters in an inner node in this loss signal. We therefore present a novel loss function, our second contribution, which specifically punishes the later, i.e., wrong line parameters in inner nodes. We do so by formulating three additional losses which we combine with cross-entropy into a new SegForestNet-specific loss function. Each new loss is designed to punish a specific undesirable trait of the region map $\textbf{R}$.

First, as an intermediate result used later by two of the new losses, we compute
\begin{equation}
    \textbf{Y}_\textbf{B}^i = \sum_{(x,y) \in \textbf{B}} \textbf{Y}_{yx} \cdot \textbf{R}_{iyx}
    \label{eq:yblockregion}
\end{equation}
for every $8\times 8$ pixel block $\textbf{B}$ and region/leaf node $i$ where every $\textbf{Y}_{yx}$ is a one-hot class vector. $\textbf{Y}_\textbf{B}^i$ is therefore a weighted vector containing the number of pixels per class in region $i$ of block $\textbf{B}$. This vector is weighted by $\textbf{R}$, i.e., when a pixel belongs only partially to a region $i$ according to $\textbf{R}$ it is counted only partially for $\textbf{Y}_\textbf{B}^i$. From $\textbf{Y}_\textbf{B}^i$ the size $s_\textbf{B}^i$ of a region in pixels and the region's class probability distribution $\textbf{P}_\textbf{B}^i$ can be computed easily:
\begin{equation}
    s_\textbf{B}^i = \sum_{c=1}^{|C|} \textbf{Y}_\textbf{B}^i(c)
\end{equation}
\begin{equation}
    \textbf{P}_\textbf{B}^i = \frac{\textbf{Y}_\textbf{B}^i}{s_\textbf{B}^i}
\end{equation}
where $\textbf{Y}_\textbf{B}^i(c)$ is the number of pixels in region $i$ of block $\textbf{B}$ belonging to class $c$.

Our first new loss specifically punishes region maps $\textbf{R}$ which define segmentations with regions $i$ in blocks $\textbf{B}$ which contain more than one class according to the ground truth. We calculate
\begin{equation}
    \mathcal{L}_\textbf{Y} = \frac{1}{N} \sum_{i,\textbf{B}} H(\textbf{P}_\textbf{B}^i)
    \label{eq:lossY}
\end{equation}
with the Gini impurity $H(\textbf{P}_\textbf{B}^i) = 1 - \sum_{c=1}^{|C|} \left(\textbf{P}_\textbf{B}^i(c)\right)^2$ and $N$ equal to the number of blocks $\textbf{B}$ multiplied by the number of leaf nodes $i$ per BSP tree. $\mathcal{L}_\textbf{Y}$ is the average Gini impurity across all regions $i$ and blocks $\textbf{B}$. We use $H$ as a proxy for the entropy. Using the actual entropy instead did not produce better results, however, due to including a logarithm, it was slower and less numerically stable to calculate. $\mathcal{L}_\textbf{Y}$ ensures that every predicted region $i$ contains only a single class according to the ground truth as this loss becomes smaller the closer $\textbf{P}_\textbf{B}^i$ is to a one-hot vector, i.e., only one class is in the partition $i$ of block $\textbf{B}$.

We also compute
\begin{equation}
    \mathcal{L}_s = \frac{1}{N} \sum_{i,\textbf{B}} \max\{s_\text{min} - s_\textbf{B}^i, 0\}
    \label{eq:minregionsizeloss}
\end{equation}
with a hyperparameter $s_\text{min}$ specifying a minimum desired region size. A minimum region size ensures that the predicted lines used for partitioning intersect the blocks they belong to. This is done to make sure that the model quickly learns to utilize all available regions rather then trying to find a solution that utilizes only a subset of the regions which may result in a larger error long-term but may present itself as an undesirable local minimum during training. We observed that most $8\times 8$ blocks contain only a single class according to the ground truth. Therefore, using only a single or very few partitions/regions per block may present itself as a local minimum to the model that is hard to avoid or to get out of during training.

While we do not enforce constrains on the normal $\textbf{n}$ and the distance $d$ used to calculate the signed distance function $f$ in Eq. \ref{eq:nosigmoid}, we still want our model to favor predictions which result in sharp boundaries between regions $i$, rather than having lots of pixels which belong to multiple regions partially. The hyperparameter $\lambda$ can address this issue in theory, however, for any adjustments made to $\lambda$, the model may learn to adjust its predictions of $\textbf{n}$ and $d$ accordingly, rendering $\lambda$ useless. We therefore also compute the loss
\begin{equation}
    \mathcal{L}_\textbf{R} = \frac{1}{|\textbf{I}|} \sum_{(x,y) \in \textbf{I}} H(\textbf{R}_{yx})
    \label{eq:lossR}
\end{equation}
where $\textbf{I}$ is the set of image pixels and $\textbf{R}_{yx}$ is the probability vector defining how likely pixel $(x, y) \in \textbf{I}$ belongs to any of the regions $i$. By minimizing $H$, we make sure that our model favors predictions of $\textbf{R}_{yx}$ which are one-hot, thus defining sharp boundaries between regions. Again, the Gini impurity $H$ serves as a proxy to the entropy producing results of equal quality.

The full loss function we use for model training consists of cross-entropy $\mathcal{L}_\text{CE}$ and the three additional region map $\textbf{R}$ specific components just described:
\begin{equation}
    \mathcal{L}_\text{total} = \mu_1 \mathcal{L}_\text{CE} + \mu_2 \mathcal{L}_\textbf{Y} + \mu_3 \mathcal{L}_s + \mu_4 \mathcal{L}_\textbf{R}.
    \label{eq:totalloss}
\end{equation}
We set the loss weights $\mu_i$ s.t. $\sum_{i=1}^4 \mu_i = 1$. This constraint only limits the range for hyperparameter optimization as any desired effective $\mu_i$ can be achieved by adjusting the constrained $\mu_i$ together with the learning rate. $\mathcal{L}_\text{CE}$ is the most important component of the loss as it is the only component affecting the leaf nodes and therefore the predicted class logits. All four loss components affect the region map $\textbf{R}$ and therefore the inner nodes and the line parameters they contain. As initially described, $\mathcal{L}_\textbf{Y}$, $\mathcal{L}_s$ and $\mathcal{L}_\textbf{R}$ are specifically designed to improve the predicted $\textbf{R}$.

\subsection{Class-Specific Trees per Block}

In order to allow our model, SegForestNet to learn class-specific partitionings of $8 \times 8$ blocks into regions, we allow partitioning the classes $C$ into subsets $C_j$ and predicting a separate BSP tree for each subset. By creating a single subset with all classes our model can emulate the original BSPSegNet. By creating subsets which contain exactly one class we can create class-specific trees. Our approach generalizes to any partitioning of $C$ though.

For each subset $C_j$ we create a separate pair of decoders (shape and content). The encoder's output is split among all subsets s.t. there is no overlap in features going into any pair of two distinct decoders. The output dimension of the content decoder for each $C_j$ is set to $|C_j|\cdot N_\text{leaves}$ where $N_\text{leaves}$ is the number of leaf nodes per tree, i.e., each content decoder only predicts class logits for the classes in its respective $C_j$. For each $C_j$ we obtain a semantic segmentation with a reduced number of classes. To obtain the final semantic segmentation with all classes we simply concatenate the class logits predicted by all the trees.

The computation of $\mathcal{L}_\text{total}$ so far only considered a single predicted tree. However, all loss components but $\mathcal{L}_\text{CE}$ need to account for the fact that there are now multiple predicted trees. First, we calculate a separate $\textbf{Y}_\textbf{B}^i$ (Eq. \ref{eq:yblockregion}) for each $C_j$. To do so, we split the one-hot vector $\textbf{Y}_{yx}$ into $\textbf{Y}_{yx}^j$, which consists of all the classes in $C_j$, and $\textbf{Y}_{yx}^{\lnot j}$, which consists of all the other classes. From these two vectors we compute a new one-hot class vector
$\textbf{Y}_{yx}^{j'}$ by appending $\sum_{c \in C_{\lnot j}} \textbf{Y}_{yx}^{\lnot j}(c)$ to $\textbf{Y}_{yx}^j$. The one-hot vector $\textbf{Y}_{yx}^{j'}$ then consists of all classes in $C_j$ with an extra dimension representing all other classes. This vector can be used to compute tree-specific versions of $\textbf{Y}_\textbf{B}^i$ which can then be used for tree-specific loss components $\mathcal{L}_\textbf{Y}^j$, $\mathcal{L}_s^j$ and $\mathcal{L}_\textbf{R}^j$. Finally, we use the equations
\begin{equation}
    \mathcal{L}_\textbf{Y} = \sum_j \frac{|C_j|}{|C|} \cdot \mathcal{L}_\textbf{Y}^j
\end{equation}
\begin{equation}
    \mathcal{L}_s = \sum_j \frac{|C_j|}{|C|} \cdot \mathcal{L}_s^j
\end{equation}
\begin{equation}
    \mathcal{L}_\textbf{R} = \sum_j \frac{|C_j|}{|C|} \cdot \mathcal{L}_\textbf{R}^j
\end{equation}
to calculate the loss components necessary for Eq. \ref{eq:totalloss}, i.e., we weight each tree-specific loss by the number of classes in its respective subset $C_j$.

Since $F_{tree}$ (Fig. \ref{fig:decoder}) becomes very small very quickly when creating partitionings of $C$ with small $|C_j|$ we omit the constraint $F_{encoder} < F_{tree}$ when predicting class-specific trees for each block. We still set $F_{encoder}$ to small values, though, to stay in line with the original idea of creating a bottleneck.

\subsection{Training Processes}
\label{sec:trainingprocessoverview}

The two training processes we compare in the next section for our fourth contribution mostly share the same underlying approach but use different hyperparameters.

\begin{figure}
    \centering
    \includegraphics[width=.5\textwidth]{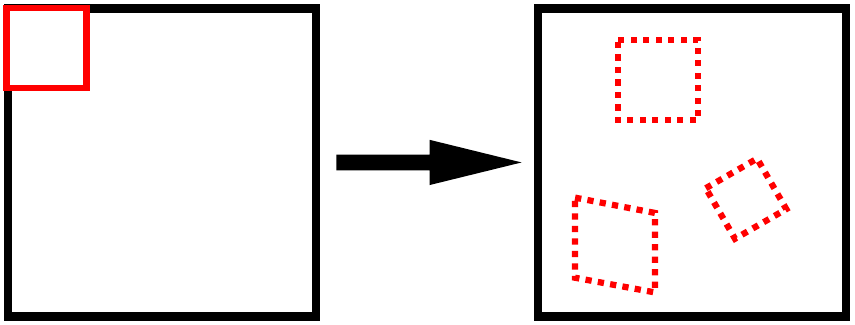}
    \caption{Sampling from a random training image (black outline). After creating a square in the top-left corner, a random affine transformation is applied to the square. The transformed square (dashed red outline) is the actual area being sampled. Using different affine transformations, multiple different training samples can be drawn from each image.}
    \label{fig:sampling}
\end{figure}

In our approach, we create training samples by starting with a square in the top-left corner of a random training image. The square's size is chosen s.t. it exactly covers as many pixels as should be in each training sample, e.g., $256 \times 256$ pixels. We then apply a random affine transformation to translate and rotate this square into a random position and orientation in the image. Scaling and shearing may also be applied with the only constraint being equal scaling along both dimensions in order to keep the shape a square. This is shown in Fig. \ref{fig:sampling}. The transformed square then defines the image area being sampled for the respective training sample. We use nearest neighbor interpolation for sampling from ground truth/label images and bilinear interpolation for all other image types, e.g., RGB images.

After sampling, all channels $c$ are normalized via $c' = \frac{c - \mu_c}{\sigma_c}$ with the mean $\mu_c$ and standard deviation $\sigma_c$ computed across all pixels of all training images. The normalization is followed by radiometric augmentation: first, all pixels are multiplied by a random contrast factor, before a random brightness offset is added. Finally, random noise drawn from $\mathcal{N}(0,1)$ multiplied by a random magnitude factor is added. While the contrast and brightness changes are equal across all channels and pixels, the noise is sampled independently for each channel and pixel. As a final augmentation, there is a chance to apply horizontal and/or vertical flipping.

For validation and test samples, we again start with a square in the top-left corner. However, instead of an affine transformation or the other previously mentioned augmentations, we simply slide this square across the image to create non-overlapping samples covering the entire image without the need for any kind of interpolation. Normalization is still applied though. Optionally, we perform test-time augmentation, i.e., neighboring samples overlap by $50\%$ and we create all eight variants that can be created by applying rotations from $\{0\degree, 90\degree\, 180\degree, 270\degree\}$ and flipping for each sample. This kind of augmentation still avoids the need for interpolation. With test-time augmentation, we get between $8$ and $32$ predictions for each pixel in each validation or test image. Therefore, for each pixel, we average the predicted logits before determining the pixel's class.

The models we trained for our evaluation usually expect RGB images, i.e., exactly three input channels. However, the training, validation and test samples we created using the approach we just described always had more than three channels: all of our datasets include near-infrared and height/depth channels in addition to RGB data. Using this information we also computed the NDVI \cite{rouse1973monitoring} as an additional channel. Therefore, we replace in the initial convolution layer of each model's encoder with a new, randomly initialized layer to expand the model's number of input channels. Even when using a pre-trained encoder we still replace this layer.

As loss we use categorical cross-entropy or a model-specific loss which includes categorical cross-entropy in the cases PFNet and SegForestNet. To account for the imbalaced classes, we use different weights per class. All weights are initially set to $1 - \frac{N_C}{N}$ where $N_C$ is the number of pixels of class $C$ in the training images and $N$ is the total number of pixels in all training images. The weight of the background class is set to $0$ in order to ignore it. Optionally, we dynamically update the class weights after every epoch with the strategy described in \cite{wittich2021appearance}: the new weights for non-ignored classes are $\left(1-\left(\text{IoU}_C - \text{mIoU}\right)\right)^\kappa$ with $\kappa = 4$ (as per \cite{wittich2021appearance}), $\text{IoU}_C$ being the Intersection-over-Union for class $C$ and $\text{mIoU}$ being the mean Intersection-over-Union.

As optimizer we use AdamW \cite{kingma2014adam,loshchilov2017decoupled} and anneal the learning rate using a cosine schedule \cite{loshchilov2016sgdr} We do not use early stopping but rather store the model weights after every epoch and compute the mean $F_1$ score across the classes on the validation set. After the training finishes, we use this metric to identify the model weights to use for evaluation on the test set.

\section{Evaluation}
In this section we evaluate SegForestNet in comparison to multiple other state-of-the-art models using several aerial image datasets. We will compare two different training processes using different augmentation strategies to show that an optimized process is more important than a segmentation model specialized on aerial images. Also, we will show the improvements of our contributions from BSPSegNet to SegForestNet in an ablation study.

\subsection{Datasets}
We used eight different datasets for evaluation. We used aerial images, in particular true orthophotos, from the German cities Hannover, Buxtehude, Nienburg, Schleswig, Hameln, Vaihingen and Potsdam. The later two datasets are part of the ISPRS 2D Semantic Labeling Benchmark Challenge \cite{ISPRSBenchmark}. Additionally, we used images from Toulouse \cite{roscher2020semcity}. The Toulouse dataset also includes instance labels for buildings for performing panoptic segmentation or instance segmentation. However, we only used the semantic labels. 

Hannover, Buxtehude, and Nienburg consist of 16 patches of $2500\times 2500$ pixels each. We omitted one such patch from Hannover since it consisted almost entirely of trees. We randomly divided the image patches into subsets s.t. roughly $70\%$ of pixels were used for training with the remainder split evenly into validation and test sets. Schleswig and Hameln consist of $1000\times 1000$ pixel patches which were also randomly divided into training, validation and test like Hannover, Buxtehude, or Nienburg. However, we only used $50\%$ of pixels for training. Schleswig has $26$ patches and Hameln has $37$ patches. For Vaihingen and Potsdam, we used the images originally released without ground truth as test images and used the remaining images as training (roughly $90\%$ of pixels) and validation (roughly $10\%$ of pixels). Toulouse consists of $16$ patches, each with a size of $3504\times 3452$ pixels. We used the two patches defined as training set for panoptic segmentation by the dataset's authors as validation set and the two patches defined as panoptic test set as our semantic test set. The remaining $12$ patches were used for training. Some patches in Toulouse have multiple associated ground truth images with differing quality. We always chose the image with the highest label quality.

All datasets except Vaihingen, which has no green channel, consist of at least RGB images. All the German cities  also have infrared channels and a digital surface model ("height" or "depth"). Toulouse actually consists of multispectral images with eight channels, including RGB and infrared. 

The ground truth classes of the German city images are impervious surfaces, buildings, low vegetation, tree, car, and clutter/background, the later two being rare. The Toulouse dataset also contains sports venues and water as separate classes in addition to the classes used for segmentation of the German cities. In Toulouse the classes car, sports venues, water, and background are rare. Since the background class is so rare and also is used to express uncertainty about the true class in the case of at least Potsdam, we ignored this class for all metrics, including training losses. The class distributions and ground sampling distances of each dataset we used are shown in Table \ref{tab:classdist}.

\begin{table}
    \centering
    \caption{Class distributions and ground sampling distances of the datasets used for evaluation. Each dataset is identified by its first letter except for Hameln (Hm). A dash indicates that the dataset's class definition does not include the respective class.}
    \label{tab:classdist}
    \begin{tabular}{r|cccccccc}
         & H & B & N & S & Hm & V & P & T \\
         \hline
         size [MPixels] & $100$ & $100$ & $100$ & $26$ & $37$ & $168$ & $1368$ & $194$ \\
         GSD [cm] & $20$ & $20$ & $20$ & $20$ & $20$ & $8$ & $5$ & $50$ \\
         imp. surface [\%] & $33.4$ & $23.6$ & $17.4$ & $14.1$ & $18.8$ & $27.8$ & $29.6$ & $24.2$ \\
         building [\%] & $39.0$ & $19.6$ & $18.2$ & $14.7$ & $19.1$ & $26.0$ & $25.7$ & $22.4$ \\
         low vegetation [\%] & $12.4$ & $36.4$ & $48.3$ & $38.9$ & $36.3$ & $21.3$ & $22.6$ & $26.9$ \\
         tree [\%] & $13.3$ & $17.7$ & $14.7$ & $31.5$ & $24.5$ & $22.9$ & $15.5$ & $16.2$ \\
         car [\%] & $1.7$ & $1.1$ & $0.6$ & $0.8$ & $1.3$ & $1.2$ & $1.8$ & $2.0$ \\
         water [\%] & - & - & - & - & - & - & - & $3.7$ \\
         sports venue [\%] & - & - & - & - & - & - & - & $1.4$ \\
         background [\%] & $0.2$ & $1.6$ & $0.8$ & $0$ & $0$ & $0.8$ & $4.8$ & $3.2$
    \end{tabular}
\end{table}

\subsection{Training Process Variants}
\label{sec:trainingprocessvariants}

In subsection \ref{sec:trainingprocessoverview} we described the model training process we used for our evaluation. We trained the models in two different variants: process variant $P_G$ is a continuation of our own approach from \cite{gritzner2021semantic}, while the variant $P_W$ is heavily inspired by the approach used in \cite{wittich2021appearance}. The hyperparameters used in each variant are shown in Tables \ref{tab:processhyperparameters} and \ref{tab:learningrates}. The random translation was chosen s.t. the square being sampled from was entirely within the image assuming a scaling of $1$, no rotation, and no shearing. The hyperparameters for both processes were optimized using random searches with search intervals initially close to the values used in the papers the processes were inspired by.

A significant difference is using the same $8000$ training samples again and again for every epoch in $P_G$, but using a set of $2500$ newly created samples for every epoch in $P_W$. Other notable differences between the two processes lie in the geometric augmentation, especially shearing and flipping, more radiometric augmentation that just adding random noise, updating the class weights after every epoch and using an encoder pre-trained on ImageNet \cite{ILSVRC15,rw2019timm}. In preliminary experiments we noticed that using pre-trained weights speeds up convergence but does not result in a better performance in the end. In $P_G$, each epoch lasts for more than twice as many iterations and each iteration includes $50\%$ more training samples (mini-batch size), therefore even the models which trained for fewer epochs in $P_G$ overall trained for longer than in $P_W$. We document this last difference for completeness but it does not affect the results. The decoder of each model was always randomly initialized.

\begin{table}
    \centering
    \caption{The hyperparameters used for the two training process variants we evaluated. Model-specific hyperparameters can be found in Table \ref{tab:learningrates}.}
    \label{tab:processhyperparameters}
    \begin{tabular}{r|cc}
    & $P_G$ & $P_W$ \\
    \hline
    pre-trained encoder & \xmark & \cmark \\
    epochs & model-specific & $105$ \\
    samples per epoch & $8000$ & $2500$ \\
    new samples each epoch & \xmark & \cmark \\
    sample size [pixels] & $224\times 224$ & $256\times 256$ \\
    mini-batch size & $18$ & $12$ \\
    $\beta_1$ (optimizer) & $0.9$ & $0.75$ \\
    $\beta_2$ (optimizer) & $0.999$ & $0.999$ \\
    weight decay (optimizer) & $0.01$ & $0.0078$ \\
    max. gradient $L_2$ norm & $\infty$ & $2$ \\
    dynamic class weights & \xmark & \cmark \\
    scaling & $\mathcal{U}(0.7, 1.3)$ & $\mathcal{U}(0.9, 1.1)$ \\
    rotation & $\mathcal{U}(-30\degree, 30\degree)$ & $\mathcal{U}(-180\degree, 180\degree)$ \\
    hor. shearing & $\mathcal{U}(-6\degree, 6\degree)$ & 0 \\
    vert. shearing & $\mathcal{U}(-6\degree, 6\degree)$ & 0 \\
    chance of hor. flipping & $0\%$ & $50\%$ \\
    chance of vert. flipping & $0\%$ & $50\%$ \\
    contrast factor & $1$ & $\mathcal{U}(0.9, 1.1)$ \\
    brightness offset & $0$ & $\mathcal{U}(-0.1, 0.1)$ \\
    noise magnitude & $\mathcal{U}(0, 0.1)$ & $\mathcal{U}(0, 0.1)$ \\
    test-time augmentation & \xmark & \cmark
    \end{tabular}
\end{table}

\begin{table}
    \begin{center}
    \caption{The model-specific initial learning rates used from which the cosine annealing started. For $P_G$ the number of training epochs is also model-specific and shown here.}
    \label{tab:learningrates}
    \begin{tabular}{r|ccc}
        & epochs ($P_G$) & learn. rate ($P_G$) & learn. rate ($P_W$) \\
        \hline
        FCN & $120$ & $0.006$ & $0.001$ \\
        SegForestNet & $200$ & $0.003$ & $0.0025$ \\
        DeepLab v3+ & $80$ & $0.0035$ & $0.001$ \\
        PFNet & $100$ & $0.005$ & $0.001$ \\
        FarSeg & $120$ & $0.001$ & $0.001$ \\
        U-Net & $80$ & $0.00015$ & $0.0005$ \\
        RA-FCN & $120$ & $0.0002$ & $0.0005$ 
    \end{tabular}
    \end{center}
\end{table}

\subsection{Test Setup}

We used PyTorch v1.13.1 with CUDA 11.6 to train models on nVidia GeForce RTX 3090 GPUs. For hyperparameter optimization we trained models with random configurations on Hannover, Vaihingen, and Toulouse for a mixture of different ground sampling distances. After every epoch during training we computed the mean $F_1$ score ($F_1$ score averaged across classes) on the respective validation sets and then used the maximum of these mean $F_1$ scores across all epochs to determine the performance of a given hyperparameter configuration. With the optimized hyperparameters we trained each combination of model and dataset ten times and computed means and standard deviations on the respective test set from these runs for our comparative evaluation in subsection \ref{sec:modelcomparison}. For our ablation study we only used five runs per configuration. We used the process variant $P_W$ for SegForestNet hyperparameter optimization and in our ablation study.

In our comparative evaluation we tested a mixture of general segmentation models and models specifically optimized for aerial images according the authors as shown in Table \ref{tab:modelsize}. All models except U-Net are defined in terms of an interchangeable encoder and a specific decoder. We used Xception \cite{chollet2017xception} as encoder as we already did in \cite{gritzner2021semantic}. We used U-Net \cite{ronneberger2015u} in the exact architecture as defined in the original paper.

\begin{table}
\caption{The models used in our evaluation. Note: SegForestNet's size is dependant on the number of trees predicted for each block. These numbers are for a configuration in which one BSP tree per block and class was predicted, therefore making the actual size dependant on the number of classes in the dataset.}
\label{tab:modelsize}
\begin{center}
\begin{tabular}{r|cc}
 & model parameters [M] & aerial image optimized \\
\hline
FCN \cite{long2015fully} & $20.9$ & \xmark \\
DeepLab v3+ \cite{chen2018encoder} & $22.2$ & \xmark \\
SegForestNet & $22.3 - 22.8$ & \cmark \\
PFNet \cite{li2021pointflow} & $26.4$ & \cmark \\
FarSeg \cite{zheng2020foreground} & $29.2$ & \cmark \\
U-Net \cite{ronneberger2015u} & $31.4$ & \xmark \\
(S-)RA-FCN \cite{mou2019relation} & $40.0$ & \cmark
\end{tabular}
\end{center}
\end{table}

We also made some small modifications to either improve the models' performances slightly or simplify them while keeping their performance the same: we use LeakyReLUs \cite{xu2015empirical} instead of ReLUs and we removed all atrous/dilated convolutions. For DeepLab v3+ we modified the strides of the encoder s.t. the spatial bottleneck had a downsampling factor of $8$.

In the case of SegForestNet we also used an encoder downsampling factor of $8$, resulting in predicting BSP trees for each $8\times 8$ block of the image. For the decoders we used $8$ residual blocks with $F_{intermediate} = 96$ (see Fig. \ref{fig:decoder}). For the shape features, $F_{encoder}$ and $F_{tree}$ were set to $8$ and $9$ respectively. For the content features, we chose $F_{encoder}=24$ and $F_{tree}=4\cdot |C|$ with $|C|$ being the number of classes in the dataset. The numbers for $F_{tree}$ are the result of using BSP trees with depth $=2$, i.e., trees with three inner nodes and four leaf nodes. In the comparative evaluation, we configured SegForestNet to predict one BSP tree per block and class, i.e., we partitioned the set classes $C$ s.t. each partition contained exactly one class. Therfore, in this case $F_{tree} = 4$.


\subsection{SegForestNet Hyperparameters}

We introduced several new hyperparameters throughout our contributions, $\lambda$ in Eq. \ref{eq:nosigmoid} and $\mu_1$ to $\mu_4$ in Eq. \ref{eq:totalloss}. As we expect $\lambda$ to have no real effect anyway, especially since we introduced $\mathcal{L}_\textbf{R}$ (Eq. \ref{eq:lossR}), we simply set $\lambda = 1$ and did not study it further. We used random search to optimize the different $\mu_i$. We set $s_\text{min} = 8$ (Eq. \ref{eq:minregionsizeloss}), i.e., each of the four regions should use at least one eighth of the area of each block. We chose this as a compromise, so that each region in a $8\times 8$ block contributes a significant part without restricting the model too much by setting $s_\text{min}$ too high.

\begin{figure*}
    \centering
    \includegraphics[width=\textwidth]{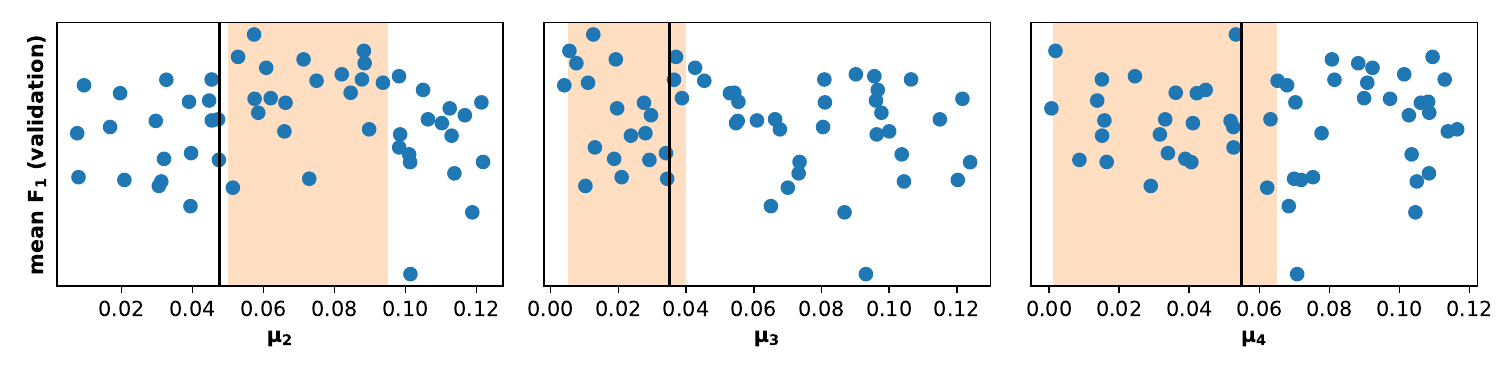}
    \caption{Validation mean $F_1$ score on Toulouse across $\mu_i$ (Eq. \ref{eq:totalloss}). The highlighted intervals show the optimal range for each $\mu_i$ (based on the highest and lowest values in the interval) and the vertical bar the value that was actually chosen. The chosen $\mu_2$ is not optimal, however, the results on the other two datasets (Hannover and Vaihingen; not shown) justify this choice, which is still near optimal for Toulouse.}
    \label{fig:hyperparameters}
\end{figure*}

An excerpt of the results of our random search is shown in Fig. \ref{fig:hyperparameters}. The impact of cross-entropy, which was controlled by $\mu_1$, was the most significant. Due to the constraint $\sum_i \mu_i = 1$ the value for $\mu_1$ is defined implicitly. With the optimal values chosen for the other three $\mu_i$, $\mu_1$ was set to $0.8625$. The other $\mu_i$ produced a slight increase in performance but should be set to rather low values ($\mu_2 = 0.0475, \mu_3 = 0.035,$ and $\mu_4 = 0.055$). We used three different datasets in our random search in order to obtain optimized hyperparameters which generalize across datasets.

\begin{figure*}
    \centering
    \includegraphics[width=.8\textwidth]{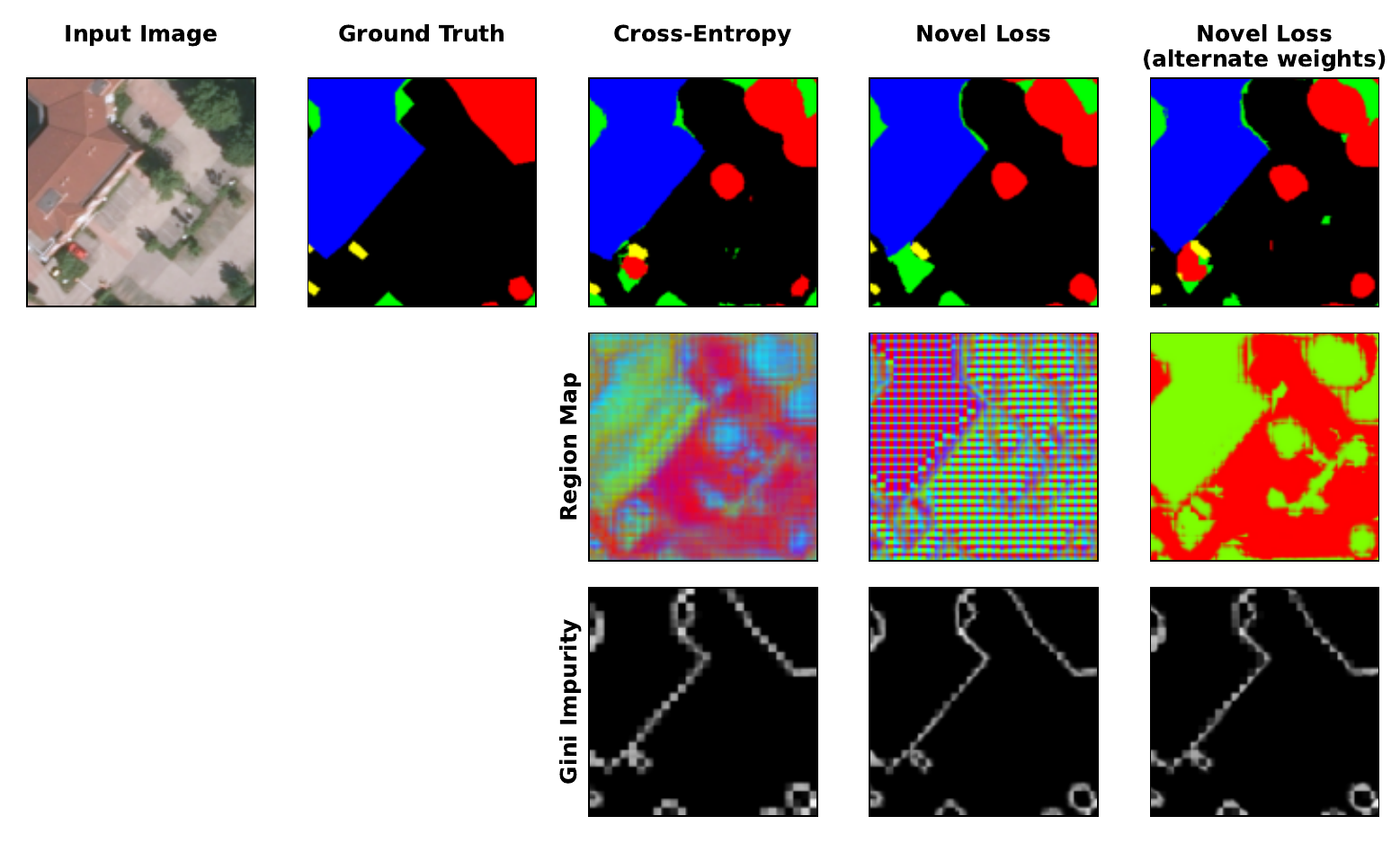}
    \caption{Visualization of the region map $\textbf{R}$ and the Gini impurity $H$ (Eq. \ref{eq:lossY}) for different loss functions and different values for $\mu_i$ (Eq. \ref{eq:totalloss}). The second to last column used optimized values for $\mu_i$, while the last column used $\mu_1 = 0.95$, $\mu_2=\mu_3=0$, and $\mu_4=0.05$. These visualizations show that our novel loss achieves the desired effects of sharp region boundaries and minimum region sizes (middle row) and pure regions containing just a single class according to the ground truth (bottom row).}
    \label{fig:regionmapvis}
\end{figure*}

In order to verify that the different loss components have the desired effects, we created a visualization of the region map $\textbf{R}$ and the Gini impurity $H$ (Eq. \ref{eq:lossY}). To create these visualizations, we used a variant of Eq. \ref{eq:finalprediction} in which we used colors or $H(\textbf{P}_\textbf{B}^i)$ in place of the class logits $\textbf{v}_i$. Using colors (red, green, blue, and cyan) instead of class logits created a visualization showing which pixel belongs to which of the four regions in a block (Fig. \ref{fig:regionmapvis}, middle row). Using $H(\textbf{P}_\textbf{B}^i)$ instead created a visualization highlighting pixels in gray or even white which belong to regions which contain more than one class according to the ground truth (Fig. \ref{fig:regionmapvis}, bottom row). Using our novel loss with optimized values for $\mu_i$ created region maps with the desired minimum region size (Eq. \ref{eq:minregionsizeloss}) and sharp region boundaries (Eq. \ref{eq:lossR}). Furthermore, the average $H(\textbf{P}_\textbf{B}^i)$ (Eq. \ref{eq:lossY}) are smaller compared to using cross-entropy. Cross-entropy also created blurry boundaries and seemed to associate certain regions with certain classes: the green region was used for buildings (blue in the ground truth), the cyan region was used for trees (red in the ground truth) and the red region for the remaining classes. Using our novel loss but with unoptimized $\mu_i$, in particular setting $\mu_2$ (pure regions containing only one class according to the ground truth; Eq. \ref{eq:lossY}) and $\mu_3$ (minimum region size; Eq. \ref{eq:minregionsizeloss}) to $0$, showed that the individual losses have the desired effects. The average $H(\textbf{P}_\textbf{B}^i)$ is slightly higher (hard to see in Fig. \ref{fig:regionmapvis}) and certain regions (blue and cyan) are never used. Still, the region boundaries are sharp since $\mu_4$ was sufficiently high.

\subsection{SegForestNet Ablation Study}

In an ablation study, shown in Table \ref{tab:ablation}, we examined the effect of each contribution individually. Each contribution individually increases the performance compared to BSPSegNet. Improved gradient flow (first column) and the novel loss (second column) have a significantly bigger impact than predicting multiple trees per block (third column). While the first two contributions on their own are already responsible for most of the performance gain, combining all three contributions is necessary to reach the best possible performance. Overall, there was an improvement of $2.8\%$ mean $F_1$ score when using all contributions. While the third contribution made the model slightly larger by using more decoders depending on the number of partitions of the set of classes, the new decoder architecture reduced the impact on the total model size due to each decoder using only $85.3$k parameters instead of $137.7$k parameters.

\begin{table}
\caption{Ablation study. The top-most row is SegForestNet with all our contributions, while the bottom-most row is the original BSPSegNet. The last column shows the mean $F_1$ score on the validation set of Buxtehude.}
\label{tab:ablation}
\begin{center}
\begin{tabular}{ccc|c}
new decoder and & novel & one BSP tree per & mean \\
region map computation & loss & block and class & $F_1$ \\
\hline
\cmark & \cmark & \cmark & $89.4\% \pm 0.1\%$ \\
\cmark & \cmark & & $89.2\% \pm 0.2\%$ \\
\cmark & & \cmark & $89.3\% \pm 0.2\%$ \\
\cmark & & & $89.3\% \pm 0.1\%$ \\
\hline
 & \cmark & \cmark & $89.3\% \pm 0.1\%$ \\
 & \cmark & & $89.2\% \pm 0.1\%$ \\
 & & \cmark & $87.0\% \pm 0.7\%$ \\
 & & & $86.6\% \pm 0.1\%$ \\
\end{tabular}
\end{center}
\end{table}

\subsection{Model Comparison}
\label{sec:modelcomparison}

\begin{figure}
    \begin{center}
    \includegraphics[width=\textwidth]{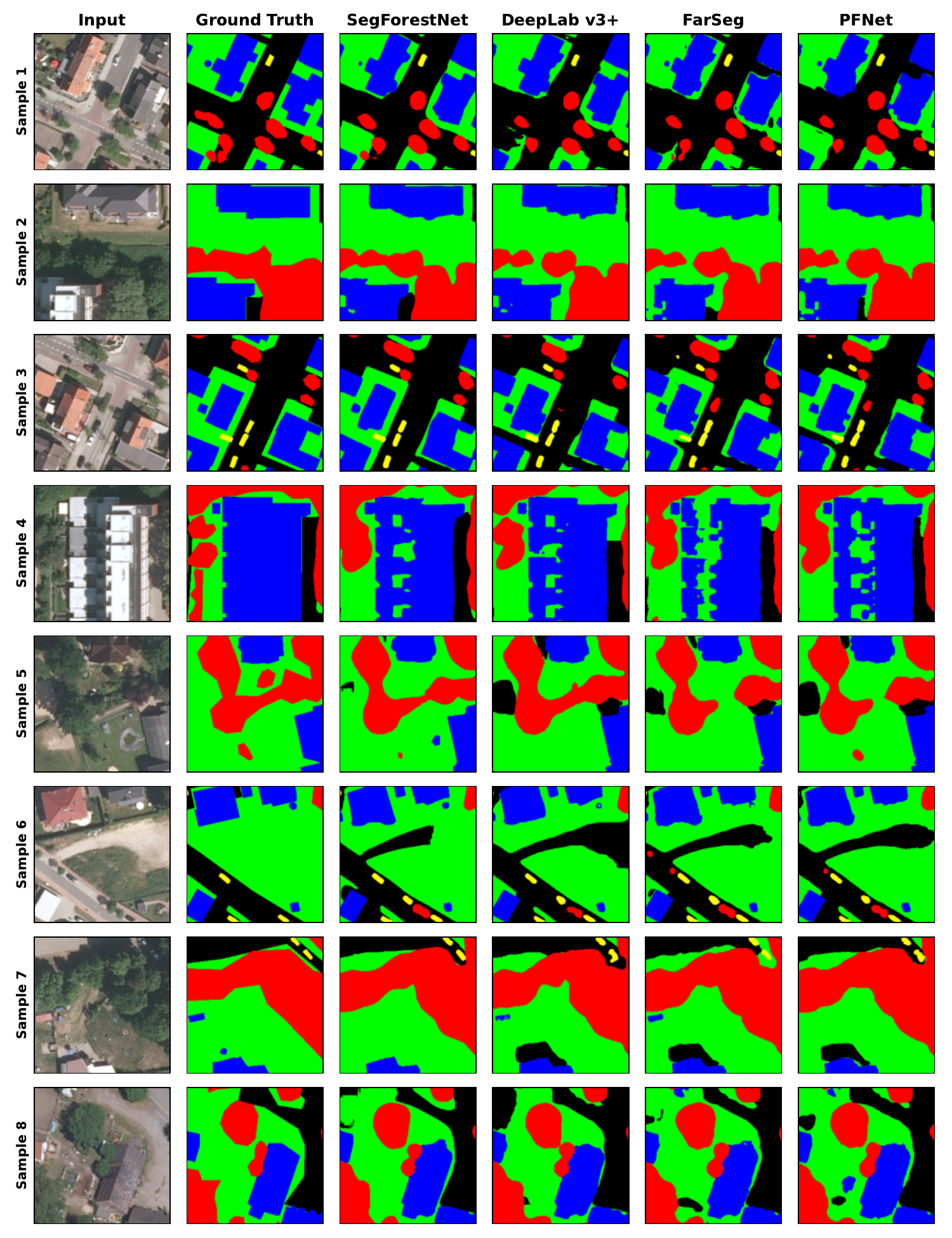}
    \includegraphics[width=.7\textwidth]{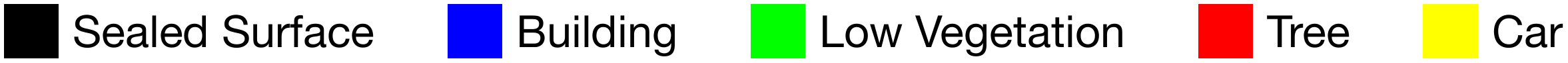}
    \end{center}
    \caption{Various samples of the predictions made by the top models on Buxtehude using the better training process variant $P_W$.}
    \label{fig:samples}
\end{figure}

In this section we evaluate the impact of the two training process variants $P_G$ and $P_W$ from subsection \ref{sec:trainingprocessvariants} on different state-of-the-art models. Results are shown in Tables \ref{tab:mf1} to \ref{tab:mf1delta} with samples of predicted segmentations in Fig. \ref{fig:samples}. Some older models, e.g., U-Net, still show competitive performance and thus we chose a selection of older and newer models as well as models optimized for segmentation in different domains such as medical images (e.g., U-Net) or aerial images (e.g., our own SegForestNet).

\begin{table}
    \caption{Mean $F_1$ score [\%] across eight datasets using process variant $P_W$. Standard deviations are shown in Table \ref{tab:mf1std}. Best model in each column shown in bold.}
    \label{tab:mf1}
    \begin{center}
    \begin{tabular}{r|cccccccc}
& H & B & N & S & Hm & V & P & T \\
\hline
FCN & 84.9 & 87.7 & 85.5 & 82.6 & 87.8 & 86.6 & 91.3 & 75.8 \\
DeepLab v3+ & \textbf{85.7} & 88.7 & 86.7 & \textbf{83.6} & 88.6 & \textbf{86.9} & \textbf{91.5} & \textbf{77.6} \\
SegForestNet & 85.5 & \textbf{88.8} & 86.2 & 83.0 & \textbf{88.7} & 86.8 & 91.3 & 74.8 \\
PFNet & 85.4 & 88.4 & 86.3 & 83.2 & 88.4 & 86.8 & \textbf{91.5} & 75.8 \\
FarSeg & \textbf{85.7} & 88.5 & \textbf{86.8} & 82.8 & 88.4 & 86.7 & 91.4 & 75.0 \\
U-Net & 84.3 & 86.7 & 85.5 & 78.5 & 86.8 & 84.2 & 88.6 & 75.2 \\
RA-FCN & 78.5 & 83.1 & 80.0 & 74.6 & 83.9 & 82.6 & 86.6 & 66.9
    \end{tabular}
    \end{center}
\end{table}

\begin{table}
    \caption{Standard deviation [\%] of the $F_1$ scores from Table \ref{tab:mf1}.}
    \label{tab:mf1std}
    \begin{center}
    \begin{tabular}{r|cccccccc}
& H & B & N & S & Hm & V & P & T \\
\hline
FCN & 0.1 & 0.1 & 0.2 & 0.6 & 0.2 & 0.1 & 0.0 & 1.9 \\
DeepLab v3+ & 0.1 & 0.1 & 0.2 & 0.8 & 0.2 & 0.1 & 0.1 & 1.6 \\
SegForestNet & 0.1 & 0.1 & 0.2 & 0.9 & 0.1 & 0.1 & 0.1 & 0.9 \\
PFNet & 0.1 & 0.2 & 0.3 & 1.2 & 0.2 & 0.1 & 0.1 & 1.6 \\
FarSeg & 0.1 & 0.2 & 0.2 & 0.9 & 0.4 & 0.1 & 0.1 & 1.2 \\
U-Net & 0.2 & 0.2 & 0.4 & 0.8 & 0.3 & 0.3 & 0.3 & 0.6 \\
RA-FCN & 1.7 & 1.4 & 2.0 & 1.5 & 1.1 & 1.2 & 1.7 & 2.4
    \end{tabular}
    \end{center}
\end{table}

\begin{table}
    \caption{Improvement in the mean $F_1$ score [\%] when going from process variant $P_G$ to $P_W$. FarSeg and U-Net are outliers. We assume U-Net would have benefitted from longer training (more epochs).}
    \label{tab:mf1delta}
    \begin{center}
    \begin{tabular}{r|cccccccc}
& H & B & N & S & Hm & V & P & T \\
\hline
FCN & 1.2 & 1.4 & 0.8 & 2.6 & 0.5 & 1.8 & 3.3 & 3.7 \\
DeepLab v3+ & 1.7 & 1.9 & 1.4 & 5.1 & 1.4 & 2.1 & 4.2 & 4.4 \\
SegForestNet & 1.1 & 1.6 & 0.7 & 4.5 & 1.0 & 1.7 & 3.5 & 0.8 \\
PFNet & 1.3 & 1.2 & 0.9 & 2.8 & 1.0 & 1.8 & 3.4 & 4.3 \\
FarSeg & 15.8 & 9.2 & 15.7 & 10.0 & 1.5 & 16.2 & 19.0 & 4.7 \\
U-Net & -0.5 & -0.8 & -0.3 & -1.2 & -0.5 & -0.3 & 1.1 & 0.2 \\
RA-FCN & -2.1 & 0.4 & -0.3 & 0.8 & 0.2 & 1.1 & 1.5 & 1.4
    \end{tabular}
    \end{center}
\end{table}

Table \ref{tab:mf1} shows that even the models not optimized for aerial images, i.e., FCN, DeepLab v3+, and U-Net, deliver competitive performance. DeepLab v3+ is even the best model in five out of the eight datasets. However, some of the aerial image optimized models such as our own SegForestNet, FarSeg and PFNet usually deliver a performance close to DeepLab v3+ and even manage to beat it on some datasets. When looking at the performance improvements provided by going from process variant $P_G$ to $P_W$ (Table \ref{tab:mf1delta}) several things are noticable. Without the better training process, FarSeg often fails to converge to good parameters resulting in massive improvements in many instances when going from $P_G$ to $P_W$. Aside from the outlier FarSeg, DeepLab v3+ has the highest performance gains. Therefore, trained with $P_G$, DeepLab v3+ and FarSeg would fall behind the other models, instead of being among the best models. SegForestNet and PFNet would be the top models indicating that optimization for aerial images actually improves aerial image segmentation performance. When comparing the standard deviations (Table \ref{tab:mf1std}) to the performance differences between the models, it becomes questionable if the best model on each dataset really outperforms the others or if the top models actually perform equally well. We therefore conclude that it is more important to have a good training process ($P_W$), which almost always improved the model performance, than to optimize specifically for aerial images. The optimization for aerial images does not hurt performance but is only really helpful when using inferior training approaches ($P_G$), i.e., in an environment when performance is not optimal anyway. Additional metrics, including the absolute performance under $P_G$ and the mIoU can be found in \ref{app:additionalmetrics}.

Fig. \ref{fig:samples} shows several samples of predictions made by the top models, SegForestNet, DeepLab v3+, FarSeg, and PFNet. The predictions are of high quality across all models. Errors are often caused by shadows, e.g, in the tree line in sample 2, near the center column in sample 4, or to the center-left in sample 5. All the models confuse the dirt road in sample 6 with a sealed surface to varying degrees. Smaller objects may be missed by some models, e.g., the small building to the center-left in sample 7. SegForestNet appears to be slightly better at predicting cars as actual rectangular shapes, whereas they may be too rounded in the predictions of other models as well as several cars merging into a single region despite being disconnected in the ground truth (samples 3, 6, and 7). These samples support the fact, that the top models are basically equally good even though their errors vary slightly.

\section{Conclusion}

In this paper we present a model for the semantic segmentation of aerial images using binary space partitioning trees, as well as three modifications to this model to improve its performance. The first modification, a refined decoder and new region map computation strategy, are aimed at improving gradients during backpropagation, while the second is a novel loss function improving the shape of the predicted segments and the last modification is an extension which enables class-specific segmentations. Taking all modifications together, our model achieves state-of-the-art performance. As a fourth contribution, we show that an optimized training process is more important than optimizing model architectures for aerial images specifically. While an optimized architecture is beneficial in a non-optimal training context, this advantage disappears in an optimized training context which is necessary to reach the best performance anyway. Our model still made better predictions for small rectangular objects, e.g., cars, where other models predicted nonsensical car shapes despite achieving basically the same overall mean $F_1$ score. In the future, we want to investigate how to let our model learn by itself what the optimal type of tree, signed distance function in each inner node, and number of trees is for a given dataset, i.e., what the optimal partitioning of classes is. This would reduce the number of design decisions having to be made when applying our model. Additionally, we want to  expand our model to be able to predict instance IDs in order to perform instance and/or panoptic segmentation.

\section*{Acknowledgments}

The aerial images of the Hannover, Buxtehude, Nienburg, Hameln, and Schleswig datasets were provided by the Landesamt für Geoinformation und Landesvermessung Niedersachsen (LGLN) and the
Landesamt für Vermessung und Geoinformation Schleswig Holstein.

\bibliographystyle{elsarticle-num} 
\bibliography{segforestnet2024}

\newpage

\begin{wrapfigure}[13]{l}{1in}
        \includegraphics[width=1in]{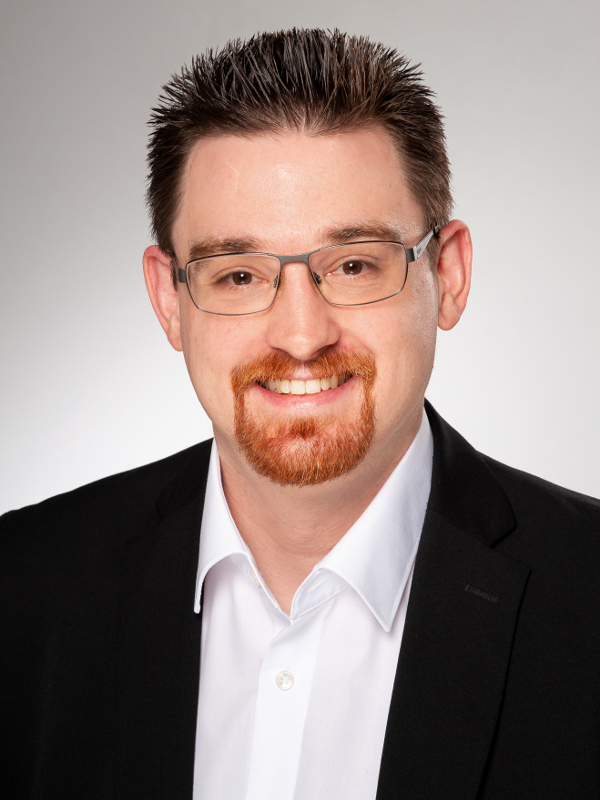}
        \includegraphics[width=1in]{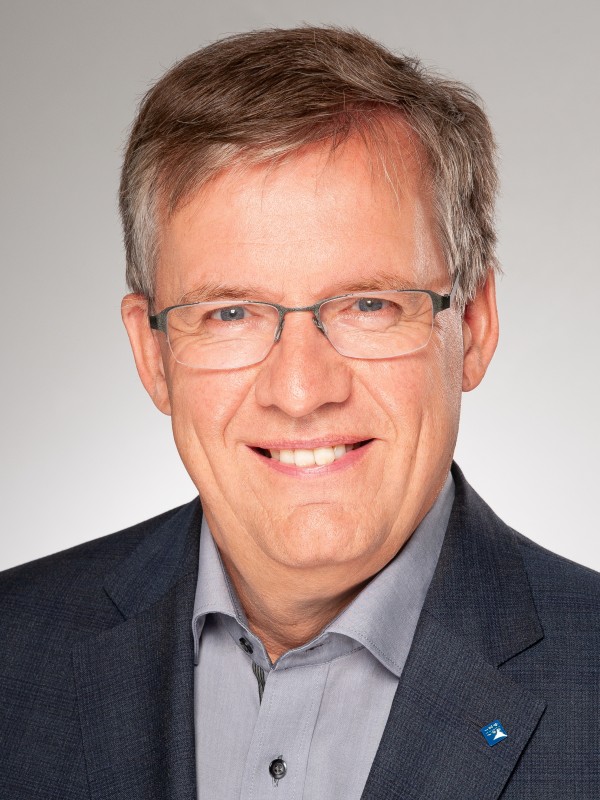}
\end{wrapfigure}
\noindent Daniel Gritzner studied Mathematics and Computer Science, with a focus on the later, at the University of Mannheim. He got his diploma and his B.Sc. in CS in 2014 and 2010 respectively. He is working on research at the intersection of computer vision and remote sensing, in particular the analysis of aerial and satellite images. His research interests are Deep Learning, Transfer Learning, and Semantic Segmentation and Object Detection for Remote Sensing Applications.

\vspace{5mm}

\noindent Prof. Dr.-Ing. Jörn Ostermann studied Electrical Engineering and Communications Engineering at the University of Hannover and Imperial College London, respectively. In 1994 he received a Dr.-Ing. from the University of Hannover for his work on low bit-rate and object-based analysis-synthesis video coding. Since 2003 he is Full Professor and Head of the Institut für Informationsverarbeitung at Leibniz Universität Hannover, Germany. In July 2020 he was appointed Convenor of MPEG Technical Coordination. He is a Fellow of the IEEE (class of 2005) and member of the IEEE Technical Committee on Multimedia Signal Processing and past chair of the IEEE CAS Visual Signal Processing and Communications (VSPC) Technical Committee. He is named as inventor on more than 30 patents. His current research interests are video coding and streaming, computer vision, machine learning, 3D modeling, face animation, and computer-human interfaces.

\newpage

\appendix

\setcounter{figure}{0} 
\setcounter{table}{0}

\section{Generalized Differentiable Rendering of Trees}
\label{app:generalization}

In this appendix we show how to generalize the
approach of using a region map $\textbf{R}$ (described in section \ref{sec:method}) for differentiable rendering generalizes to other signed distance functions $f$ and tree structures.

\subsection{Signed Distance Functions}

Eq. \ref{eq:nosigmoid} can be used with any signed distance function $f$, i.e., with any function which satisfies the following conditions:
\begin{enumerate}
    \item $f(\textbf{p}) = 0 \Leftrightarrow$ the point $\textbf{p}$ lies exactly on a boundary defined by $f$
    \item $\text{sgn}(f(\textbf{p}))$ specifies on which side of the boundary defined by $f$ the point $\textbf{p}$ lies
    \item $|f(\textbf{p})|$ is the distance of $\textbf{p}$ to the boundary defined by $f$
\end{enumerate}
The third condition can even be relaxed for the use in Eq. \ref{eq:nosigmoid}. It is sufficient if $|f(\textbf{p})|$ increases monotonically with the distance of $\textbf{p}$ to the boundary instead of it being the actual distance. Especially if $\textbf{p}$ is far from the boundary an approximate distance suffices.

\begin{figure*}
    \centering
    \includegraphics[width=.8\textwidth]{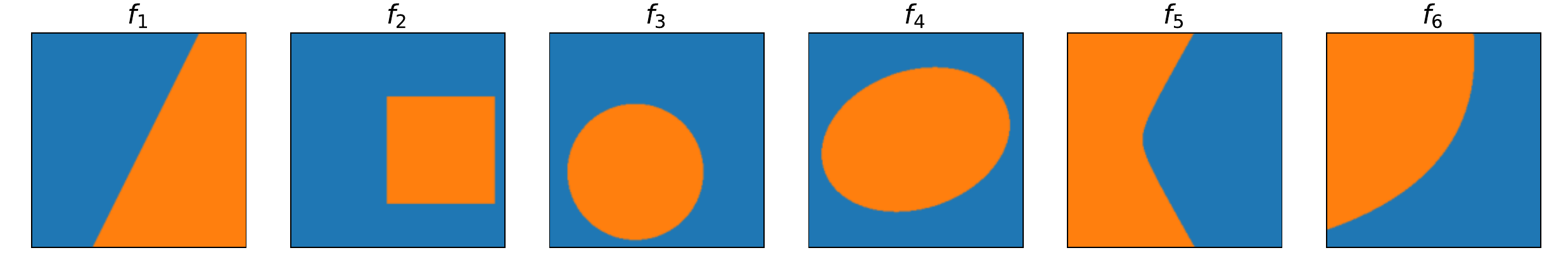}
    \caption{The decision boundaries created by signed distance functions based on different geometric primitives. From left to right: line, square, circle, ellipse, hyperbola, parabola. The blue area shows points for which the respective signed distance function is non-negative. As an example, in the inside of the circle $f_3$ is negative whereas it is positive on the outside.}
    \label{fig:sdfs}
\end{figure*}

So far, we only used lines as partition boundaries. Therefore, three values need to be predicted for each inner node: two values for the normal vector $\textbf{n}$ and one value for the distance $d$ to the origin. The signed distance $f_1(\textbf{p})$ of a point $\textbf{p}$ to the line can then be computed as
\begin{equation}
    f_1(\textbf{p}) = \textbf{n}\cdot\textbf{p} - d.
    \label{eq:linesdf}
\end{equation}
Fig. \ref{fig:sdfs} shows partitionings defined by signed distance functions based on other geometric primitives. The corresponding equations are as follows. To compute the approximate signed distance $f_2(\textbf{p})$ to a square, its center $\textbf{x}$ (two values) and its size $s$ (one value) need to be predicted:
\begin{equation}
    f_2(\textbf{p}) = \max\{|\textbf{x}_1 - \textbf{p}_1|,|\textbf{x}_2 - \textbf{p}_2|\} - s.
\end{equation}
In this equation $\textbf{x}_i$ and $\textbf{p}_i$ are the two components of the vectors $\textbf{x}$ and $\textbf{p}$. As with the square, the signed distance $f_3(\textbf{p})$ to a circle requires a predicted center $\textbf{x}$ and radius $r$ (one value) as well:
\begin{equation}
    f_3(\textbf{p}) = \lVert \textbf{x} - \textbf{p} \rVert_2^2 - r.
    \label{eq:circlesdf}
\end{equation}
This formulation is a pseudo signed distance which avoids the need to compute a square root and expects the model to learn to predict the square of the radius. The boundary of an ellipse can be defined by $d_1 + d_2 = c$ where $d_i$ are distances to fixed points and $c \in \mathbb{R}^+$ is a constant. To compute an approximate signed distance $f_4(\textbf{p})$, two points $\textbf{x}$ and $\textbf{y}$, each consisting of two values, need to be predicted as well as one value for the constant $c$. The signed distance then is
\begin{equation}
    f_4(\textbf{p}) = \lVert \textbf{x} - \textbf{p} \rVert_2 + \lVert \textbf{y} - \textbf{p} \rVert_2 - c.
\end{equation}
Similar to an ellipse, a hyperbola can be defined by $d_1 - d_2 = c$ with $d_i$ and $c$ as before. The corresponding approximate signed distance $f_5(\textbf{p})$ is
\begin{equation}
    f_5(\textbf{p}) = \left|\lVert \textbf{x} - \textbf{p} \rVert_2 - \lVert \textbf{y} - \textbf{p} \rVert_2\right| - c.
\end{equation}
A parabola can be defined as the set points satisfying $d_1 = d_2$ where $d_1$ is the distance to a fixed point and $d_2$ is the distance to a line. Therefore, five values need to be predicted, two for a fixed point $\textbf{x}$, two for a normal vector $\textbf{n}$ and one for a distance $d$ of the line to the origin. The approximate signed distance $f_6(\textbf{p})$ is
\begin{equation}
    f_6(\textbf{p}) = \lVert \textbf{x} - \textbf{p} \rVert_2 - f_1(\textbf{p})
\end{equation}
All (pseudo) signed distance functions $f_i$ except $f_2$ use formulations s.t. that arbitrarily rotated, scaled and translated instances of the underlying geometric primitives used for partitioning can be predicted.

\begin{figure}
    \centering
    \includegraphics[width=.6\columnwidth]{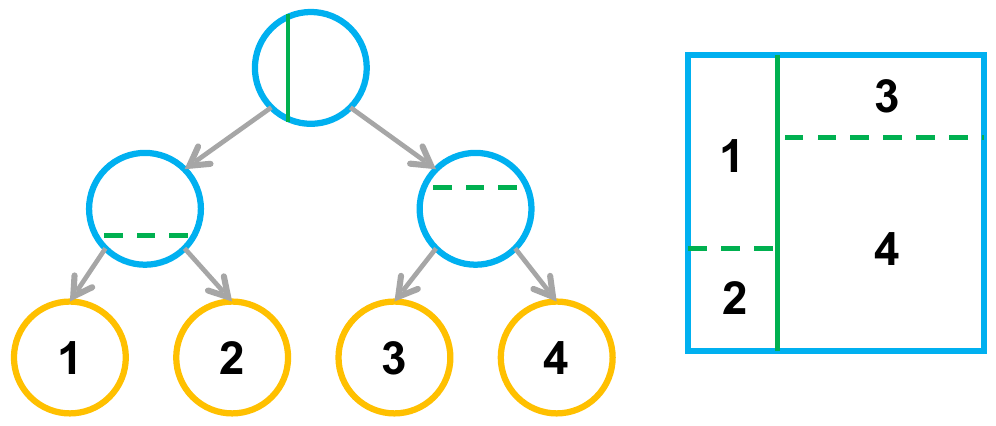}
    \caption{Visualization of a $k$-d tree (left) and a region it partitions (right). The parameters of each inner node (blue) are a fixed dimension, indicated by the orientation of the line used for partitioning, and a predicted threshold, indicated by the position of the line along the fixed dimension.}
    \label{fig:kdtree}
\end{figure}

\subsection{Tree Structure}
While BSPSegNet and SegForestNet only use BSP trees, differentiable rendering of other tree structures is possible by using the same region map $\textbf{R}$ approach. A $k$-d tree \cite{bentley1975multidimensional}, as shown in Fig. \ref{fig:kdtree}, is spatial partitioning data structure similar to a BSP tree that can be used to partition a space with an arbitrary number of dimensions. It is also a binary tree. Instead of using a signed distance function to decide on which side of a boundary a given point lies, each inner node of $k$-d tree uses a threshold and a dimension index to decide whether a point lies in the left or right subset. The threshold $t$ is a value that needs to predicted while the dimension index $i$ is a fixed value chosen when instantiating the model. Eqs. \ref{eq:nosigmoid}-\ref{eq:rright2} can be used for $k$-d trees as well with the signed distance function
\begin{equation}
    f_7(\textbf{p}) = t - \textbf{p}_i
\end{equation}
where $\textbf{p}_i$ is the $i$-th component of the vector $\textbf{p}$. An inner node of a $k$-d tree requires only a single parameter to be predicted as opposed to the three to five parameters required for the signed distance functions used by an inner node of a BSP tree. However, generally $k$-d trees need to be deeper, e.g., to be able to define slopes, which counteracts this advantage. This depth disadvantage can be mitigated by making $i$ a predicted parameter. We call a tree with such inner nodes a dynamic $k$-d tree. A downside of this mitigation strategy is that one or two additional parameters (depending on the implementation) need to be predicted per inner node to decide whether the first dimension $i=1$ or the second dimension $i=2$ shall be used for partitioning. Even with this extra flexibility a dynamic $k$-d tree still has a depth disadvantage in the worst case, e.g., in the aforementioned slope example.

\begin{figure}
    \centering
    \includegraphics[width=.6\columnwidth]{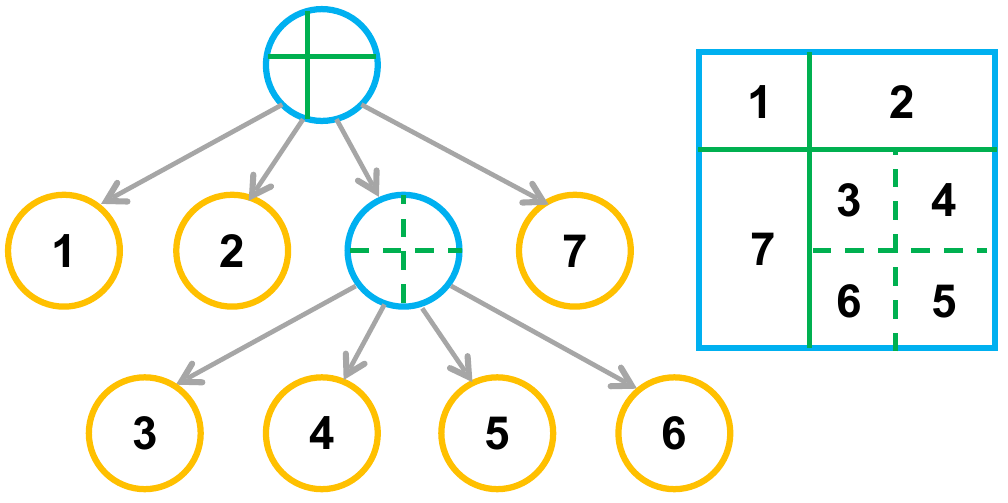}
    \caption{Visualization of a quadtree (left) and a region it partitions (right). The parameters of each inner node (blue) are the predicted coordinates of a point. This point partitions a given region into four quadrants which are then children of the associated inner node.}
    \label{fig:quadtree}
\end{figure}

A quadtree \cite{finkel1974quad}, shown in Fig. \ref{fig:quadtree}, is a spatial partitioning data structure specifically designed to partition two dimensional planes. In each inner node of a quadtree a point $\textbf{x}$ is used to partition the space into four quadrants. Compared to a $k$-d tree, a quadtree inner node partitions a space along both dimensions instead of only a single fixed dimension $i$ and it uses a different threshold $t$ for each dimension by using the components of $\textbf{x}$ as thresholds. The equations
\begin{equation}
    t_1 = \textbf{x}_1 - \textbf{p}_1, t_2 = \textbf{x}_2 - \textbf{p}_2
\end{equation}
\begin{equation}
    t_3 = \text{ReLU}(t_1), t_4 = \text{ReLU}(-t_1)
\end{equation}
\begin{equation}
    t_5 = \text{ReLU}(t_2), t_6 = \text{ReLU}(-t_2)
\end{equation}
\begin{equation}
    \textbf{R}_{iyx} := \textbf{R}_{iyx} + \lambda \cdot t_4 \cdot t_5
    \label{eq:quadtreetl}
\end{equation}
\begin{equation}
    \textbf{R}_{iyx} := \textbf{R}_{iyx} + \lambda \cdot t_3 \cdot t_5
    \label{eq:quadtreetr}
\end{equation}
\begin{equation}
    \textbf{R}_{iyx} := \textbf{R}_{iyx} + \lambda \cdot t_4 \cdot t_6
    \label{eq:quadtreebl}
\end{equation}
\begin{equation}
    \textbf{R}_{iyx} := \textbf{R}_{iyx} + \lambda \cdot t_3 \cdot t_6
    \label{eq:quadtreebr}
\end{equation}
are used to update the region map $\textbf{R}$ for a quadtree at a location $\textbf{p} = (x,y)$. Note: in Eqs. \ref{eq:quadtreetl}-\ref{eq:quadtreebr} $i$ refers to indices of leaf nodes reachable from one of the child nodes. In Eq. \ref{eq:quadtreetl} $i$ needs to be set to the indices of the leaf nodes covering the top-left partition, in Eq. \ref{eq:quadtreetr} $i$ refers to the top-right partition, in Eq. \ref{eq:quadtreebl} $i$ refers to the bottom-left partition and in Eq. \ref{eq:quadtreebr} $i$ refers to the bottom-right partition.

This region map $\textbf{R}$ based framework is general enough to even allow for trees in which different inner nodes are instances of different tree types (BSP, $k$-d, quadtree) or use different signed distance functions in the case of BSP tree inner nodes. When using class-specific trees, i.e., when partitioning the set of classes $C$ into subsets $C_j$, separate region maps specific to each of the subsets have to be computed anyway. Consequently, different tree configurations can be used for each of the subsets $C_j$ if so desired.

\section{Additional Evaluation Metrics}
\label{app:additionalmetrics}

Tables \ref{tab:oldmf1} and \ref{tab:oldmf1std} show the performance of the models when trained using process variant $P_G$. SegForestNet, PFNet, and U-Net become the top models, leaving DeepLab v3+ and FarSeg, models which show strong performance under $P_W$, behind. FarSeg especially has a very high standard deviation, indicating that it often fails to converge properly when using $P_G$ instead of $P_W$. Note: PFNet and especially SegForestNet are much smaller than U-Net in terms of number of parameters, yet still deliver performance on a similar level as U-Net or even manage to beat it sometimes. This indicates that the aerial image optimization in the former two models are helpful inductive biases. Tables \ref{tab:miou} to \ref{tab:oldmioustd} show the model performances in mIoU instead of the mean $F_1$ score.

\begin{table}
    \caption{Mean $F_1$ score [\%] across eight datasets using process variant $P_G$. Standard deviations are shown in Table \ref{tab:oldmf1std}. Best model in each column shown in bold.}
    \label{tab:oldmf1}
    \begin{center}
    \begin{tabular}{r|cccccccc}
& H & B & N & S & Hm & V & P & T \\
\hline
FCN & 83.7 & 86.3 & 84.7 & 80.0 & 87.4 & 84.8 & 88.0 & 72.1 \\
DeepLab v3+ & 83.9 & 86.8 & 85.3 & 78.5 & 87.2 & 84.8 & 87.3 & 73.2 \\
SegForestNet & 84.4 & 87.2 & 85.4 & 78.5 & \textbf{87.7} & \textbf{85.1} & 87.8 & 74.0 \\
PFNet & 84.1 & 87.3 & 85.4 & \textbf{80.4} & 87.3 & \textbf{85.1} & \textbf{88.1} & 71.5 \\
FarSeg & 70.0 & 79.3 & 71.1 & 72.8 & 86.9 & 70.5 & 72.4 & 70.3 \\
U-Net & \textbf{84.7} & \textbf{87.5} & \textbf{85.8} & 79.7 & 87.3 & 84.4 & 87.5 & \textbf{75.0} \\
RA-FCN & 80.6 & 82.6 & 80.3 & 73.9 & 83.7 & 81.5 & 85.1 & 65.5
    \end{tabular}
    \end{center}
\end{table}

\begin{table}
    \caption{Standard deviation [\%] of the $F_1$ scores from Table \ref{tab:oldmf1}.}
    \label{tab:oldmf1std}
    \begin{center}
    \begin{tabular}{r|cccccccc}
& H & B & N & S & Hm & V & P & T \\
\hline
FCN & 0.1 & 0.4 & 0.2 & 0.8 & 0.2 & 0.2 & 0.1 & 1.9 \\
DeepLab v3+ & 0.1 & 0.4 & 0.2 & 0.6 & 0.3 & 0.2 & 0.2 & 1.7 \\
SegForestNet & 0.1 & 0.3 & 0.1 & 0.7 & 0.1 & 0.2 & 0.3 & 1.8 \\
PFNet & 0.1 & 0.1 & 0.1 & 0.7 & 0.2 & 0.3 & 0.2 & 0.9 \\
FarSeg & 28.3 & 21.0 & 28.5 & 20.3 & 0.8 & 28.5 & 29.2 & 2.6 \\
U-Net & 0.1 & 0.5 & 0.4 & 0.5 & 0.2 & 0.3 & 0.2 & 0.9 \\
RA-FCN & 0.4 & 0.7 & 0.3 & 2.0 & 0.6 & 1.4 & 0.6 & 1.2
    \end{tabular}
    \end{center}
\end{table}

\begin{table}
    \caption{Mean Intersection-over-Union (mIoU) [\%] across eight datasets using process variant $P_W$. Standard deviations are shown in Table \ref{tab:mioustd}. Best model in each column shown in bold.}
    \label{tab:miou}
    \begin{center}
    \begin{tabular}{r|cccccccc}
& H & B & N & S & Hm & V & P & T \\
\hline
FCN & 74.5 & 78.6 & 75.0 & 72.6 & 78.8 & 76.7 & 84.3 & 62.3 \\
DeepLab v3+ & 75.6 & 80.0 & 76.7 & \textbf{73.7} & 79.9 & \textbf{77.1} & \textbf{84.6} & \textbf{64.1} \\
SegForestNet & 75.3 & \textbf{80.2} & 76.0 & 73.1 & \textbf{80.0} & \textbf{77.1} & 84.3 & 62.0 \\
PFNet & 75.2 & 79.6 & 76.2 & 73.3 & 79.6 & 77.0 & 84.5 & 62.1 \\
FarSeg & \textbf{75.7} & 79.7 & \textbf{76.9} & 72.9 & 79.6 & 76.8 & 84.4 & 61.4 \\
U-Net & 73.5 & 77.0 & 75.0 & 68.2 & 77.1 & 73.1 & 79.8 & 62.1 \\
RA-FCN & 65.9 & 72.4 & 68.0 & 64.7 & 73.2 & 71.0 & 76.8 & 52.4
    \end{tabular}
    \end{center}
\end{table}

\begin{table}
    \caption{Standard deviation [\%] of the mIoUs from Table \ref{tab:miou}.}
    \label{tab:mioustd}
    \begin{center}
    \begin{tabular}{r|cccccccc}
& H & B & N & S & Hm & V & P & T \\
\hline
FCN & 0.1 & 0.2 & 0.4 & 0.6 & 0.4 & 0.2 & 0.0 & 2.2 \\
DeepLab v3+ & 0.2 & 0.2 & 0.3 & 0.8 & 0.2 & 0.2 & 0.1 & 1.9 \\
SegForestNet & 0.2 & 0.2 & 0.3 & 0.9 & 0.2 & 0.2 & 0.2 & 0.8 \\
PFNet & 0.2 & 0.4 & 0.4 & 1.0 & 0.3 & 0.2 & 0.1 & 1.8 \\
FarSeg & 0.1 & 0.3 & 0.3 & 0.8 & 0.6 & 0.1 & 0.1 & 1.4 \\
U-Net & 0.2 & 0.3 & 0.6 & 0.7 & 0.4 & 0.4 & 0.5 & 0.6 \\
RA-FCN & 2.2 & 1.8 & 2.4 & 1.3 & 1.6 & 1.7 & 2.4 & 2.8
    \end{tabular}
    \end{center}
\end{table}

\begin{table}
    \caption{Improvement in the mIoU [\%] when going from process variant $P_G$ to $P_W$.}
    \label{tab:mioudelta}
    \begin{center}
    \begin{tabular}{r|cccccccc}
& H & B & N & S & Hm & V & P & T \\
\hline
FCN & 1.7 & 1.9 & 1.2 & 2.8 & 0.7 & 2.6 & 5.4 & 2.9 \\
DeepLab v3+ & 2.5 & 2.8 & 2.2 & 5.2 & 2.1 & 3.0 & 6.7 & 3.3 \\
SegForestNet & 1.6 & 2.3 & 1.1 & 4.5 & 1.5 & 2.6 & 5.7 & 1.0 \\
PFNet & 1.9 & 1.7 & 1.5 & 3.3 & 1.6 & 2.6 & 5.4 & 2.8 \\
FarSeg & 15.3 & 9.7 & 15.2 & 9.6 & 2.2 & 16.0 & 20.7 & 3.6 \\
U-Net & -0.7 & -1.2 & -0.3 & -1.4 & -0.9 & -0.4 & 1.7 & -0.7 \\
RA-FCN & -2.7 & 0.4 & -0.1 & 0.6 & 0.0 & 1.3 & 2.2 & 0.0
    \end{tabular}
    \end{center}
\end{table}

\begin{table}
    \caption{MIoU [\%] across eight datasets using process variant $P_G$. Standard deviations are shown in Table \ref{tab:oldmioustd}. Best model in each column shown in bold.}
    \label{tab:oldmiou}
    \begin{center}
    \begin{tabular}{r|cccccccc}
& H & B & N & S & Hm & V & P & T \\
\hline
FCN & 72.8 & 76.7 & 73.8 & 69.8 & 78.1 & 74.1 & 78.9 & 59.5 \\
DeepLab v3+ & 73.1 & 77.2 & 74.5 & 68.5 & 77.9 & 74.1 & 77.9 & 60.9 \\
SegForestNet & 73.8 & 77.9 & 74.8 & 68.6 & \textbf{78.6} & \textbf{74.5} & 78.6 & 61.0 \\
PFNet & 73.3 & 77.9 & 74.7 & \textbf{70.0} & 78.0 & 74.4 & \textbf{79.1} & 59.3 \\
FarSeg & 60.3 & 70.0 & 61.6 & 63.3 & 77.4 & 60.8 & 63.7 & 57.8 \\
U-Net & \textbf{74.2} & \textbf{78.2} & \textbf{75.3} & 69.5 & 78.0 & 73.5 & 78.1 & \textbf{62.8} \\
RA-FCN & 68.6 & 72.0 & 68.1 & 64.1 & 73.2 & 69.7 & 74.6 & 52.4
    \end{tabular}
    \end{center}
\end{table}

\begin{table}
    \caption{Standard deviation [\%] of the mIoUs from Table \ref{tab:oldmiou}.}
    \label{tab:oldmioustd}
    \begin{center}
    \begin{tabular}{r|cccccccc}
& H & B & N & S & Hm & V & P & T \\
\hline
FCN & 0.2 & 0.5 & 0.2 & 0.7 & 0.4 & 0.3 & 0.1 & 1.4 \\
DeepLab v3+ & 0.2 & 0.5 & 0.3 & 0.5 & 0.4 & 0.2 & 0.4 & 1.2 \\
SegForestNet & 0.2 & 0.4 & 0.1 & 0.6 & 0.1 & 0.3 & 0.4 & 1.7 \\
PFNet & 0.2 & 0.1 & 0.2 & 0.6 & 0.2 & 0.4 & 0.3 & 0.7 \\
FarSeg & 26.0 & 19.7 & 26.2 & 18.5 & 1.1 & 26.4 & 27.5 & 3.1 \\
U-Net & 0.2 & 0.7 & 0.7 & 0.5 & 0.3 & 0.4 & 0.3 & 0.7 \\
RA-FCN & 0.5 & 0.8 & 0.3 & 1.7 & 0.8 & 1.6 & 0.9 & 1.1
    \end{tabular}
    \end{center}
\end{table}






\end{document}